\documentclass{article}

\usepackage{arxiv}
\usepackage{graphicx}

\usepackage[utf8]{inputenc} 
\usepackage[T1]{fontenc}    
\usepackage{graphicx} 
\usepackage{amssymb}
\usepackage{algorithm}
\usepackage{algpseudocode}
\usepackage{amsmath}
\usepackage{natbib}
\usepackage{xcolor}
\usepackage{booktabs}
\usepackage[inkscapelatex=false]{svg}
\usepackage{subcaption}
\usepackage{hyperref}

\def\MambaSimple{Mamba-2S}
\def\AlgoName{2Mamba}
\def\AlgoNameExp{2Mamba-E}

\title{2Mamba2Furious: Linear in Complexity, Competitive in Accuracy} 

\author{
  Gabriel Mongaras \\
  Lyle School of Engineering\\
  Southern Methodist University\\
  Dallas, TX 75205 \\
  \texttt{gabriel@mongaras.com} \\
  \And
  Eric C. Larson \\
  Lyle School of Engineering\\
  Southern Methodist University\\
  Dallas, TX 75205 \\
  \texttt{eclarson@smu.edu} \\
}

\date{\today}

\begin{document}

\maketitle

\begin{abstract}
Linear attention transformers have become a strong alternative to softmax attention due to their efficiency. However, linear attention tends to be less expressive and results in reduced accuracy compared to softmax attention. To bridge the accuracy gap between softmax attention and linear attention, we manipulate Mamba-2, a very strong linear attention variant. We first simplify Mamba-2 down to its most fundamental and important components, evaluating which specific choices make it most accurate. From this simplified Mamba-2 variant (\MambaSimple{}), we improve the $A$-mask and increase the order of the hidden state, resulting in a method, which we call \AlgoName{}, that is nearly as accurate as softmax attention, yet much more memory efficient for long context lengths. We also investigate elements to Mamba-2 that help surpass softmax attention accuracy. Code is provided for all our experiments \footnote{\href{https://github.com/gmongaras/2Mamba2Furious}{https://github.com/gmongaras/2Mamba2Furious}} \footnote{\href{https://huggingface.co/collections/gmongaras/2mamba2furious-linear-in-complexity}{https://huggingface.co/collections/gmongaras/2mamba2furious-linear-in-complexity}}.
\end{abstract}
%
%
\section{Introduction}
%
%
Transformers have become the standard architectural backbone for modern language models. The core of the transformer is softmax attention, which routes information between all tokens in a set. While softmax attention is a highly accurate model component, it imposes quadratic complexity with respect to the sequence length during training. When causal, softmax attention has linear complexity during inference. Although the complexity of softmax attention cannot be reduced to linear complexity during training, algorithms such as Flash Attention (\citet{flash_attn}) make softmax attention more computationally efficient by utilizing highly efficient CUDA kernels (tiling). However, even with a highly optimized kernel, the underlying algorithm still retains quadratic complexity in FLOP count. 
\vspace{0.5em}
\\
Linear attention algorithms (\citet{linear_attn}) aim to reduce the complexity of softmax attention by replacing the exponential nonlinearity in the softmax function with a decomposable kernel function. These algorithms are linear during training and constant during inference, making them much more desirable than softmax attention, solely based on algorithmic complexity. Another advantage of linear attention is that it can be implemented as an RNN (\citet{rnns}), which increases efficiency for long sequence inference. While linear attention has linear complexity during training and constant complexity during inference, the accuracy is consistently worse than that of softmax attention. 
\vspace{0.5em}
\\
More recent works in NLP applications make naive linear attention more expressive. Two such prominent works are Mamba (\citet{mamba}), which adds semi-separable decay components to attention, and DeltaNet (\citet{deltanet}), which uses the delta rule (\citet{deltarule}), modeling the linear attention update rule as a form of gradient descent. The accuracy of these algorithms is significantly better than vanilla linear attention but still falls short of full softmax attention. Motivated by this finding, we aim to build on these efficient attention alternatives and increase model accuracy closer to that of softmax attention.
Specifically, we (1) employ Mamba-2 (\citet{mamba2}) as a base model, (2) isolate the important components of the algorithm, and (3) use a higher order hidden state to reach softmax-level accuracy while keeping the complexity of the model linear. We build upon the work of (\citet{on_the_expr_of_sm_attn}), who showed that higher order hidden states with linear attention get closer to softmax-level accuracy. Via the Taylor expansion of the exponentiated query-key inner product, one obtains softmax attention from linear attention with higher order hidden states. We leverage this finding to improve the expressivity of Mamba-2. Additionally, we show that using an exponentiated query-key inner product improves the Mamba-2 model and outperforms softmax attention, albeit at the cost of requiring a $KV$ cache. We also explore the similarity of this approach to the forgetting transformer (\citet{fox_forgetting_trans}).
\section{Background}
\subsection{Softmax Attention}

(\citet{og_attn}) introduced softmax attention and (\citet{attn_is_all_you_need}) popularized the use of softmax attention by creating the transformer architecture for natural language translation. Since then, the transformer can be found in most modern machine learning architectures and has been used in a variety of applications such as in computer vision (\citet{vit}), reinforcement learning (\citet{dec_transformer}), robotics (\citet{rt_1}), generative image models (\citet{dalle_2}) (\citet{hun_vid}) (\citet{whisper}), and medicine (\citet{alphafold}), among many other applications. The softmax attention formulation is found in equation \ref{eq:sm_attn} (without the $\frac{1}{\sqrt{d_k}}$ term for simplicity).

\begin{gather}
    Q = X W_Q \in \mathbb{R}^{H, N, d_h} \quad K = X W_K \in \mathbb{R}^{H, N, d_h} \quad V = X W_V \in \mathbb{R}^{H, N, d_h} \nonumber \\
    O = \text{softmax} \left( Q K^T + M \right) V \quad=\quad \frac{\exp \hspace{-0.2em} \left( Q K^T + M \right) V}{\sum_j \exp \hspace{-0.2em}\left( Q K^T + M \right)} \label{eq:sm_attn}
\end{gather}

The exponential term forces a quadratic FLOP count with respect to the sequence length during training and a linear FLOP count during inference, which is not ideal for long context applications. Several methods have been developed to reduce the quadratic bottleneck and make computing the quantity more efficient. Most notably, Flash Attention (\citet{flash_attn}) computes softmax attention in tiles for better GPU utilization. Although Flash Attention is more efficient than computing softmax attention in native PyTorch, the FLOP count remains quadratic.

\noindent
\begin{minipage}[c]{0.40\textwidth}
    \begin{algorithm}[H]
    \begin{algorithmic}
    \caption{Linear attention}\label{alg:lin_attn}
    \Require $X \in \mathbb{R}^{N, d}$
    \Require $W_{QKV} \in \mathbb{R}^{d, 3 \cdot (H \cdot d_h)}$
    \Require $W_{out} \in \mathbb{R}^{(H \cdot d_h), d}$
    \Ensure $\phi(\cdot) = \text{ReLU}(\cdot)$
    \State 
    \begin{align*}
    QKV &= X W_{QKV} && \in \mathbb{R}^{N, 3\cdot(H \cdot d_h)} \\
    Q,K,V &= \text{split}(QKV) && \in \mathbb{R}^{H, N, d_h} \\
    y &= ((\phi(Q) \phi(K)^T \odot M) V) 
    && \in \mathbb{R}^{H, N, d_h} \\
    N &= \sum_j (\phi(Q) \phi(K)^T \odot M) && \in \mathbb{R}^{H, N} \\
    y_n &= y / N && \in \mathbb{R}^{H, N, d_h} \\
    y'_n & =\text{vec}(y_n)_{(H, N, d_h) \longrightarrow (N, H \cdot d_h)} &&   \in \mathbb{R}^{N, H \cdot d_h} \\
    o &= y'_n W_{out} &&   \in \mathbb{R}^{N, d}
    \end{align*}
    \end{algorithmic}
    \end{algorithm}
\end{minipage}
\hfill
\begin{minipage}[c]{0.40\textwidth}
    \begin{gather}
        Q = X W_Q \quad K = X W_K \quad V = X W_V \nonumber \\
        O = \frac{\left[ \phi(Q) \phi(K)^T \right] V}{\sum_j \phi(Q) \phi(K)^T}   \label{eq:lin_attn} \\
         \hspace{1em} = \frac{\phi(Q) \left[ \phi(K)^T V \right]}{\phi(Q) \sum_j \phi(K)^T}   \label{eq:lin_attn_eff}
    \end{gather}
\end{minipage}
%
%
%

\newpage

\subsection{Linear attention (naive computation)} 
\vspace{-5pt}Linear attention is derived by taking multi-head softmax attention and replacing the softmax nonlinearity function with a kernel function $\exp(QK^T) \Rightarrow K(Q, K) = \phi(Q) \phi(K)^T$, as shown in Algorithm \ref{alg:lin_attn}. By making this substitution, linear attention can either be computed by first taking the inner product of the queries and keys (equation \eqref{eq:lin_attn}) or by first taking the inner product of the keys and values (equation \eqref{eq:lin_attn_eff}). The latter has quadratic complexity in the dimension while the former, like softmax attention, has quadratic complexity in the sequence length. One note is that adding the causal mask requires a special kernel to make causal linear attention computationally efficient, otherwise the training complexity in naive torch is similar to softmax attention as seen in Algorithm \ref{alg:lin_attn}. While linear attention is desirable due to its linear complexity, naive linear attention falls short of softmax attention performance, as shown in Figure \ref{fig:all_attn}.

\subsection{Mamba-2}
Several approaches have attempted to improve the accuracy of linear attention while retaining linear complexity, such as performer (\citet{performer}), cosformer (\citet{cosformer}), hedgehog (\citet{hedgehog}), RetNet (\citet{retnet}), gated linear (\citet{gate_lin}), and DeltaNet (\citet{deltanet}), among others. One algorithm that is both efficient and improves linear attention significantly is Mamba-2 (\citet{mamba2}). In the description below, we examine Mamba-2 purely from an architectural perspective, not from the efficiency gained from the associative scan algorithm.
\vspace{0.5em}
\\
The original Mamba (\citet{mamba}) architecture was derived from a state space model (SSM) (\citet{ssm}) (\citet{s4}), which are not expressive on their own due to having time-independent queries, keys, and values (referred to as C, and B, and x in the SSM literature). Mamba improves SSM expressiveness by making the query, key, and value matrices time-dependent. Mamba-2 further improves efficiency and expressiveness by introducing the associative scan algorithm and decay mask ($A$ mask). The resulting algorithm can be effectively formulated as linear attention with a decay mask (section \ref{sec:isolate} expands this assertion and isolates elements of Mamba-2 that are most expressive). From this perspective, Mamba-2 takes linear attention and makes it more expressive through clever architectural elements and parallelization via associativity.
\vspace{0.5em}
\\
%
%
%
The complete Mamba-2 algorithm is found in GitHub\footnote{\href{https://github.com/state-spaces/mamba}{https://github.com/state-spaces/mamba}} and has some additions not described fully in the paper (\citet{mamba2}).
That is, the Mamba-2 codebase  builds in several inductive biases. The full Mamba model with all biases included from this codebase is written out in algorithms \ref{alg:mamba2_setup} and \ref{alg:mamba2_forward}. Note that the head and sequence dimensions are arbitrarily swapped to simplify the presented algorithm, though the swaps can be inferred via the explicit shapes. To make the comparison with other algorithms easier, as noted in (\citet{mamba2}), we employ the $Q$, $K$, and $V$ notation, rather than the equivalent $C$, $B$, and $x$ notation (respectively). Additional simplifications were made such as using a single head group, however like with softmax attention, one can have head groups akin to MQA in attention (\citet{group_query_attn}).

As seen in Algorithm \ref{alg:mamba2_setup} and Algorithm \ref{alg:mamba2_forward}, Mamba-2 has numerous elements beyond just a decay mask. Many of the choices seem arbitrary or underexplored and add complexity without much investigation or motivation. We question if the added complexity is necessary for Mamba-2 to outperform vanilla linear attention. To analyze the components that make Mamba-2 superior (and propose changes), we break Mamba-2 into its components and build up a simplified algorithm, which we name \MambaSimple{}, from these singular components.

\subsection{Softmax as a Recurrent Neural Network}

Before explaining the changes to Mamba-2, it is necessary to motivate our rationale through prior work. 
(\citet{on_the_expr_of_sm_attn}) examine the accuracy discrepancy between linear attention and softmax attention and found that taking the sum of higher powers of higher order RNNs approaches the accuracy of full softmax attention. By the Taylor expansion of the exponential\footnote{We note that the Taylor expansion used is more appropriately called the Maclaurin expansion, as the expansion point is centered on $x=0$. However we adopt the same naming convention.}, softmax attention is equivalent to the sum of all nonnegative integer powers of the query-key inner product with the denominator being just a form of normalization. Taking the sum term-by-term, higher order terms produce a larger hidden state and improve downstream accuracy, approaching softmax accuracy.  
\vspace{0.5em}

\begin{algorithm}
\begin{algorithmic}
\caption{Mamba-2 Setup}\label{alg:mamba2_setup}
\State \begin{align*}
H &= \text{num heads},\quad  d_h = \text{head (inner) dim},  \quad N = \text{seq. length,} \\[-4pt]
\mathcal{U}[\cdot]&= \text{Uniform Dist.}, \qquad d_{ssm} = H*d_h \\[-4pt]
d_{conv} &= 3 \cdot d_{ssm}, \quad dt_{min} = 0.001, \quad dt_{max}=0.1, \quad l_{min}=\log(dt_{min}), \quad l_{max}=\log(dt_{max}) \\[-4pt]
A_{log} &= \log(\mathcal{U}[1, 16]) \in \mathbb{R}^{H}, \quad D = \text{ones}({H*d_h}) \in \mathbb{R}^{H \cdot d_h} \\[-4pt]
dt_{init} &= \exp\left(x \right) \in \mathbb{R}^{H} \quad \text{where } \quad x\sim\mathcal{U}\left[l_{min}, l_{max}\right] \\[-4pt]
dt_{bias} &= \text{softplus}^{-1}(dt_{init}) \quad= dt_{init} + \log(1 - \exp(-dt_{init})) \in \mathbb{R}^{H}
\end{align*}
\end{algorithmic}
\end{algorithm}

\begin{algorithm}
\begin{algorithmic}
\caption{Mamba-2 Forward Pass}\label{alg:mamba2_forward}
\Require $h \in \mathbb{R}^{N, d}, \quad W_z \in \mathbb{R}^{d, H \cdot d_h}, \quad W_{QKV} \in \mathbb{R}^{d, d_{conv}} \quad W_{dt} \in \mathbb{R}^{d, H}$
\State \ \ \ \ \ \ \ \ \ \ \  $W_{out} \in \mathbb{R}^{H \cdot d_h, d}, \quad D \in \mathbb{R}^{H \cdot d_h}, \quad A_{log} \in \mathbb{R}^{H}, \quad dt_{bias} \in \mathbb{R}^{H}$
\State \begin{align*}
Q, K, V &= \sigma(\text{conv\_1d}(h\cdot W_{QKV})) && \in \mathbb{R}^{H, N, d_h} \\[-4pt]
dt_P &= h \cdot W_{dt} &&\in \mathbb{R}^{H, N} \\[-4pt] 
Z &= h \cdot W_{z} &&\in \mathbb{R}^{H, N, d_h} \\[-4pt]
A &= -\exp(A_{log}) &&\in \mathbb{R}^H \\[-4pt]
dt &= \text{softplus}(dt_P + dt_{bias}) &&\in \mathbb{R}^{H, N} \\[-4pt]
D_{res} &= V \odot D, \qquad V_{dt} = V \odot dt &&\in \mathbb{R}^{H, N, d_h} \\[-4pt]
A^{CS} &= \text{cumsum}(A \odot dt) &&\in \mathbb{R}^{H, N} \\[-4pt]
A^{M} &= \exp(A^{CS} - (A^{CS})^T) \quad \text{such that  } A^{M}_{ij} = \exp(A^{CS}_i - A^{CS}_j)&&\in \mathbb{R}^{H, N, N} \\
M_{ij} &= \begin{cases} 1, & \text{if } i \ge j\\ 0, & \text{if } i < j \end{cases} && M \in \mathbb{R}^{H, N, N} \\
y &= (Q K^T \odot A^M \odot M) \cdot V_{dt} &&\in \mathbb{R}^{H, N, d_h} \\[-4pt]
y_D &= y + D_{res} &&\in \mathbb{R}^{H, N, d_h} \\[-4pt]
y'_D &= \text{vec}(y_D)_{(H, N, d_h) \longrightarrow (N, H \cdot d_h)} && \in \mathbb{R}^{N, H \cdot d_h} \\[-4pt]
y_N &= \text{RMSNorm}(y'_D \odot \sigma(Z)) &&\in \mathbb{R}^{N, H \cdot d_h} \\[-4pt]
out &= y_N \cdot W_{out} &&\in \mathbb{R}^{N, d}\\[-4pt]
\end{align*}
\end{algorithmic}
\end{algorithm}

While increasing the order of linear attention increases the expressiveness of the hidden state, it also increases the hidden state on the order of $d_h^p$ where $p$ is the order of the RNN and $d_h$ is the head dimension. Vanilla linear attention turns out to be a first order approximation of softmax attention. That is, linear attention uses an additive RNN of order $p=1$ which results in a hidden state of size $(d_h, d_h)$. A higher order RNN would have a hidden state of roughly dimension $(d_h^p, d_h)$, which can be reduced as explained in section \ref{sec:rnn_comp_red}.
\vspace{0.5em}
\\
Purely from a memory perspective, the hidden state of softmax attention is the $KV$ cache, which requires $2\times N\times d_h$ elements per head. The $KV$ cache grows linearly with respect to the sequence length. While most values of $p$ would result in an unreasonable hidden state size, a second order RNN, with $p=2$ would result in a hidden state of dimension $(d^2, d)$, still reasonable in memory for a sufficiently long context window. We use this intuition to improve Mamba-2 with a second order hidden state, improving the accuracy while keeping the memory requirements lower than softmax attention for a sufficiently large context window.
\vspace{0.5em}
\\
Additionally, normalization on the query-key sum requires a strictly non-negative image of the query-key inner product. A second order hidden state has an image with strictly non-negative values. Linear attention obtains positive values by applying a nonlinearity on the pre-image of the inner product of the queries and keys, thus restricting the domain of the inner product function. Squaring the query-key inner product presents no such restriction on the pre-image, while keeping the inner product image positive. As such, normalization on the queries and keys rather than on the attention output can be used, which has been shown to be both efficient and stable when using a form of online softmax attention (\citet{online_sm}) such as flash attention (\citet{flash_attn}).
\section{Isolating Mamba-2 Accuracy Gains}
\label{sec:isolate}
To investigate the accuracy gains for each element in Mamba-2, we perform several ablations. For all our ablations, we start with the base Llama 2 (\citet{llama2}) model and replace the softmax attention block with the Mamba-2 block. Our small model is about 300 million parameters. We use this model for all ablations, which helps to facilitate rapid experimentation. The medium model is about 700 million parameters and is used after ablation experiments to examine scalability of the algorithm. The maximum sequence length of all ablations is 2048 unless specified. For comparisons and reporting, we plot the test loss over $90K$ steps \footnote{Our code trains for $100K$ steps and evaluates on the test loss every $10K$ steps. However, a bug exists which does the evaluation on 1 step after each $10K$ steps, not evaluating the $100K$'th step. We do not re-evaluate all experiments as it would take months of retraining and we believe $90K\rightarrow100K$ would not lead to any insights.} and the final test loss at the end of training. Each model is trained to perform next token language modeling using the HuggingFace FineWeb dataset (CC-MAIN-2024-51 version) (\citet{fineweb}). This dataset is composed of over 15 Trillion clean and deduplicated tokens from CommonCrawl, a dataset of a crawl of webpages on the internet. As such, it is composed of various different types of data, from code, to random documents.
\begin{figure}[htbp]
    \centering
    \includegraphics[width=0.6\textwidth]{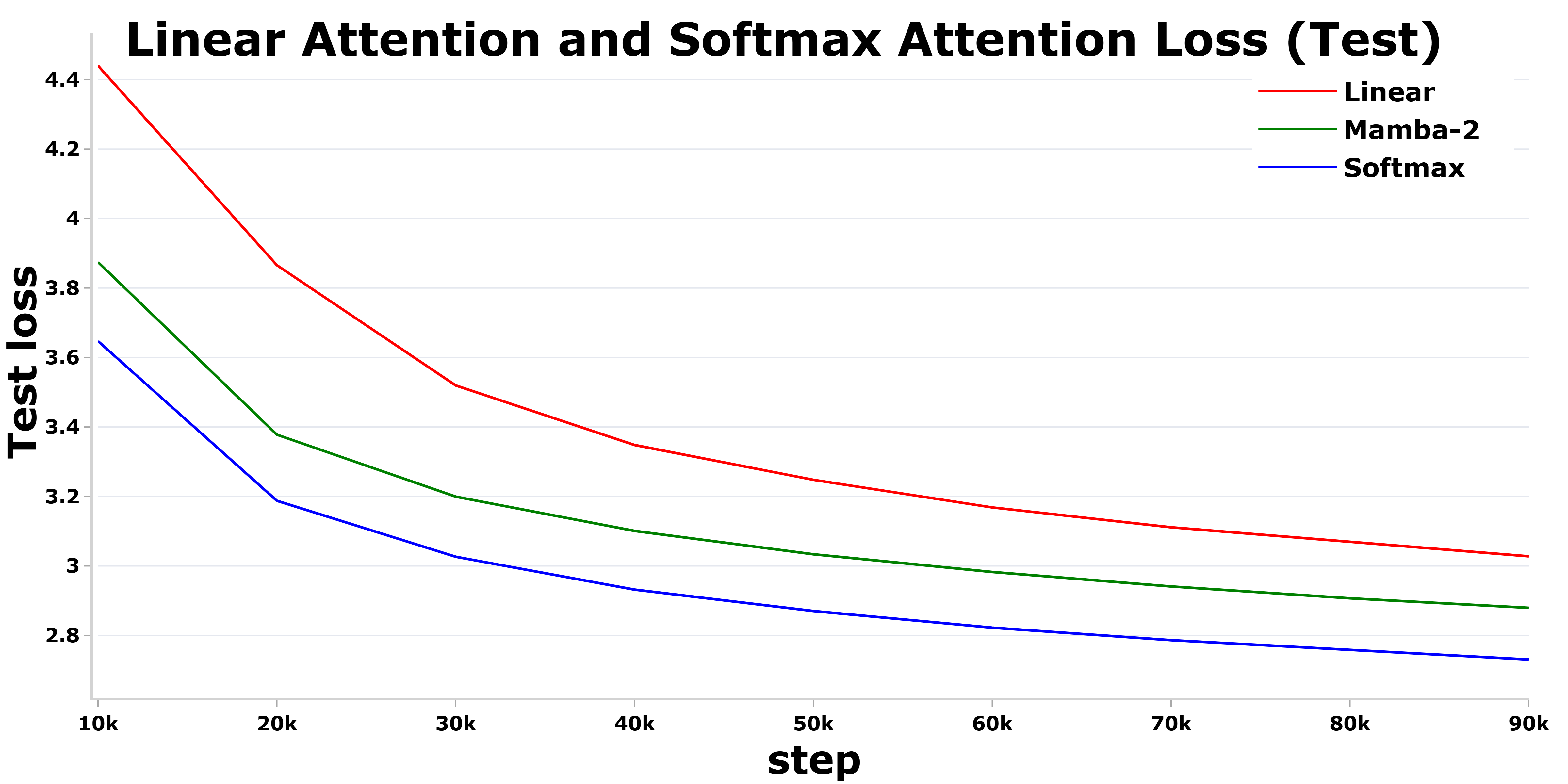}
    \caption{Accuracy of linear attention, Mamba, and softmax attention, keeping everything but the attention mechanism constant across experiments.}
    \label{fig:all_attn}
\end{figure}
Mamba-2 (\citet{mamba2}) was shown to be an expressive, powerful architecture. One that is both fast and accurate. Figure \ref{fig:all_attn} shows the test loss for Mamba-2, normal linear attention, and softmax attention. As seen in Figure \ref{fig:all_attn}, Mamba-2 is significantly better than normal linear attention and is much closer in accuracy to softmax attention. However, most elements in the codebase were not ablated, leaving to question which components are important to the expressive power of Mamba-2 and which are unnecessary. We examined the official Mamba-2 repository \footnote{\href{https://github.com/state-spaces/mamba}{https://github.com/state-spaces/mamba}} and isolate the components of the Mamba-2 block. Specifically, we ablate the following \footnote{Additional ablations found in our codebase. These are the most notable and interesting.}:
\begin{enumerate}
    \item $QK$ activation type (SiLU, ReLU or None)
    \item $A$-mask type (Original or Softplus)
    \item Input convolution window size (window size one (no convolution), two, three, or four)
    \item Additive $D$ residual (binary, present or not)
    \item Multiplicative $Z$ gate (binary, present or not)
    \item Normalization type (Output or Softmax)
    \item Value discretization $dt$ (binary, present or not)
\end{enumerate}
While most of the ablations we perform are straightforward, we want to highlight the normalization type and $A$-mask type. We call normalization on the $QK$ inner product \textit{softmax normalization}. Linear attention with a strictly positive inner product space can use softmax normalization, but Mamba-2 cannot as the queries and keys are not necessarily positive. Instead, Mamba-2 uses an RMS norm layer before the output projection. We call this form of normalization \textit{output normalization}. The $A$-mask in the original Mamba-2 paper is tied to the values by the value discretization parameter, which we call \textit{original $A$-masking}. We ablate if associating the $A$-mask with the discretization parameter is necessary. Because the $A$-mask must be strictly negative, as shown in equation \eqref{eq:A_mask_type}, we test the negative softplus function on the $A$-mask, remove the discretization parameter and name this \textit{softplus $A$-masking}. 
\vspace{0.01em}
\begin{align}
     \text{``original'' $A$-masking} & \rightarrow A = -\exp(A_{log}) \odot dt \nonumber \\
     \text{``softplus'' $A$-masking} & \rightarrow A = -\text{softplus}(A) \label{eq:A_mask_type}
\end{align}
\noindent
\begin{minipage}[c]{0.45\textwidth}
    \begin{figure}[H]
        \centering
        \includegraphics[width=\linewidth]{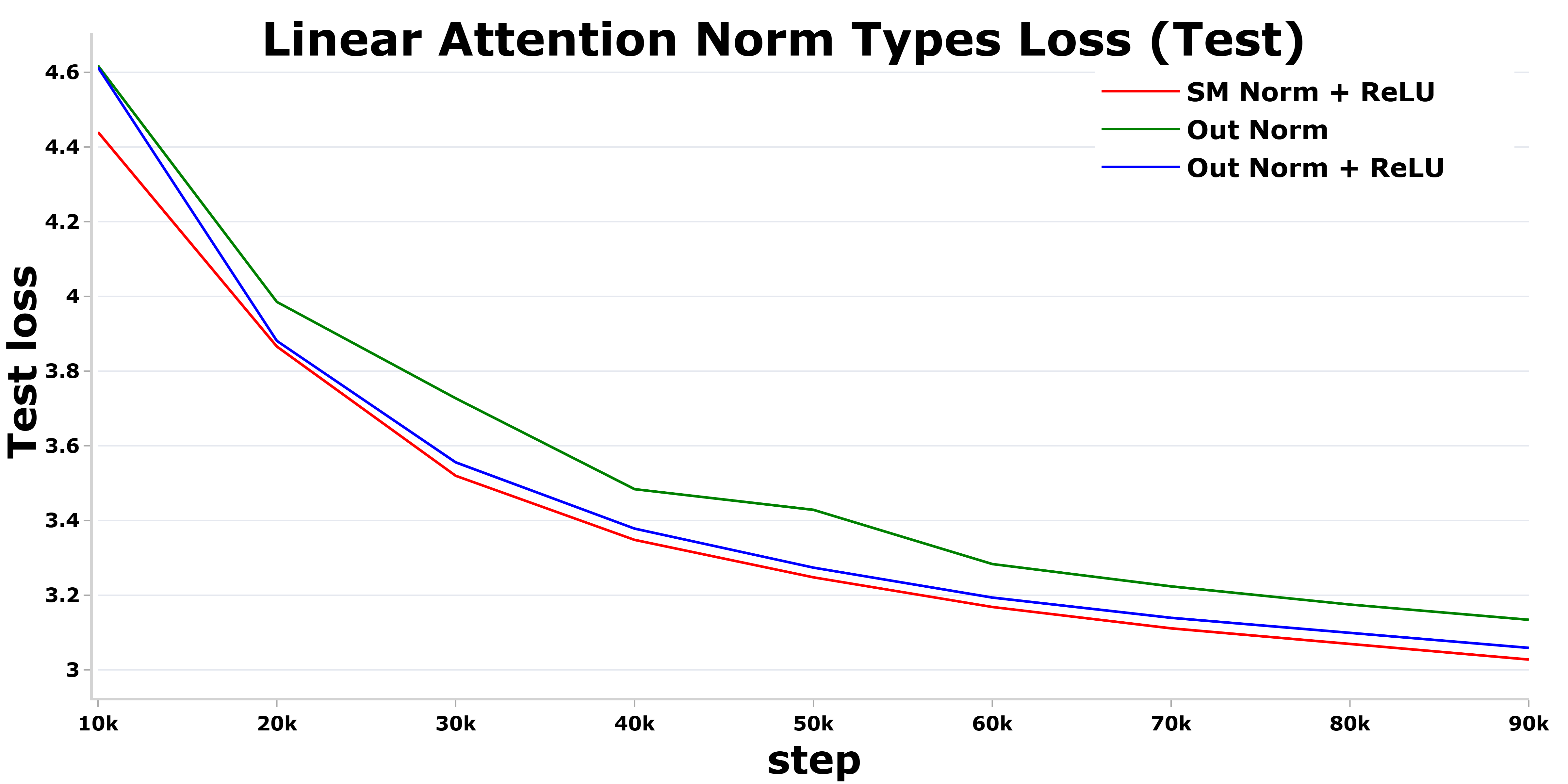}
        \caption{Accuracy of various norm types. Softmax normalization requires a positive inner-product space image, as such we use ReLU.}
        \label{fig:norm_types}
    \end{figure}
\end{minipage}
\hfill
\begin{minipage}[c]{0.52\textwidth}
    \begin{table}[H]
    \centering
    \begin{tabular}{l|cc}
      \toprule
      Model & Train loss & Test loss\\
      \midrule
      ReLU & 3.12 & 3.06 \\
      Conv (w=2) & 2.87 & 2.93 \\
      Conv (w=3) & 2.94 & 2.91 \\
      Conv (w=4) & \textbf{2.84} & 2.9 \\
      Conv (w=2) + SiLU & 2.95 & 2.9 \\
      Conv (w=3) + SiLU & 2.95 & 2.89 \\
      Conv (w=4) + SiLU & 2.9 & \textbf{2.89} \\
      \bottomrule
    \end{tabular}
    \vspace{0.25em}
    \caption{Adding a convolution to normal linear attention increases accuracy.}
    \label{tab:conv_sizes}
    \end{table}
\end{minipage}

To begin our ablation study, we need a form of linear attention that will allow for more diverse experimentation. The original linear attention formulation as proposed by (\citet{linear_attn}) has softmax-style normalization on the $QK$ inner product. This normalization restricts experimentation to a strictly non-negative inner product function range. Stability aside, as seen in Figure \ref{fig:norm_types}, adopting an output norm is nearly as accurate as softmax norm for basic linear attention. When ablating components of Mamba-2, we therefore use an output norm as it does not necessitate a strictly positive inner product space pre-image, giving more freedom to perform similar ablations. \footnote{Since linear attention with $QK$ normalization requires a strictly positive image of the $QK$ inner product, some sort of activation function must be used. As such, we cannot test \textit{SM Norm} by itself.}
\vspace{0.5em}
\\
The first element of Mamba-2 we examine is the input convolution. As seen in Table \ref{tab:conv_sizes}, adding a convolution with kernel size 2 is significantly better than normal linear attention. Increasing the convolution size to 3 or adding an activation function (SiLU used in Mamba-2) gives a slight accuracy gain, but is not as impactful as adding the convolution itself. We only test up to a convolution size of 4 as the implementation of causal\_conv\_1d\footnote{\href{https://github.com/Dao-AILab/causal-conv1d}{https://github.com/Dao-AILab/causal-conv1d}} only allows for a max window size of 4. As the window sizes increases beyond a window size of 2, the accuracy gains are small. To keep the implementation minimal, we adopt a convolution of window size 2 without an activation function. A window size of 2 minimizes the memory necessary in the complete attention operation. Increasing the convolution window requires storing $3d$ more memory per window size increase. That is, storing an additional past token for each of the queries, keys, and values. As for the activation function, the training loss appears to increase while the test loss decreases when adding the activation function. We opt to remove the activation function for simplicity, however adding it appears to slightly improve model test loss while increasing train loss.

\noindent
\begin{minipage}[c]{0.45\textwidth}
    \begin{figure}[H]
        \centering
        \includegraphics[width=\linewidth]{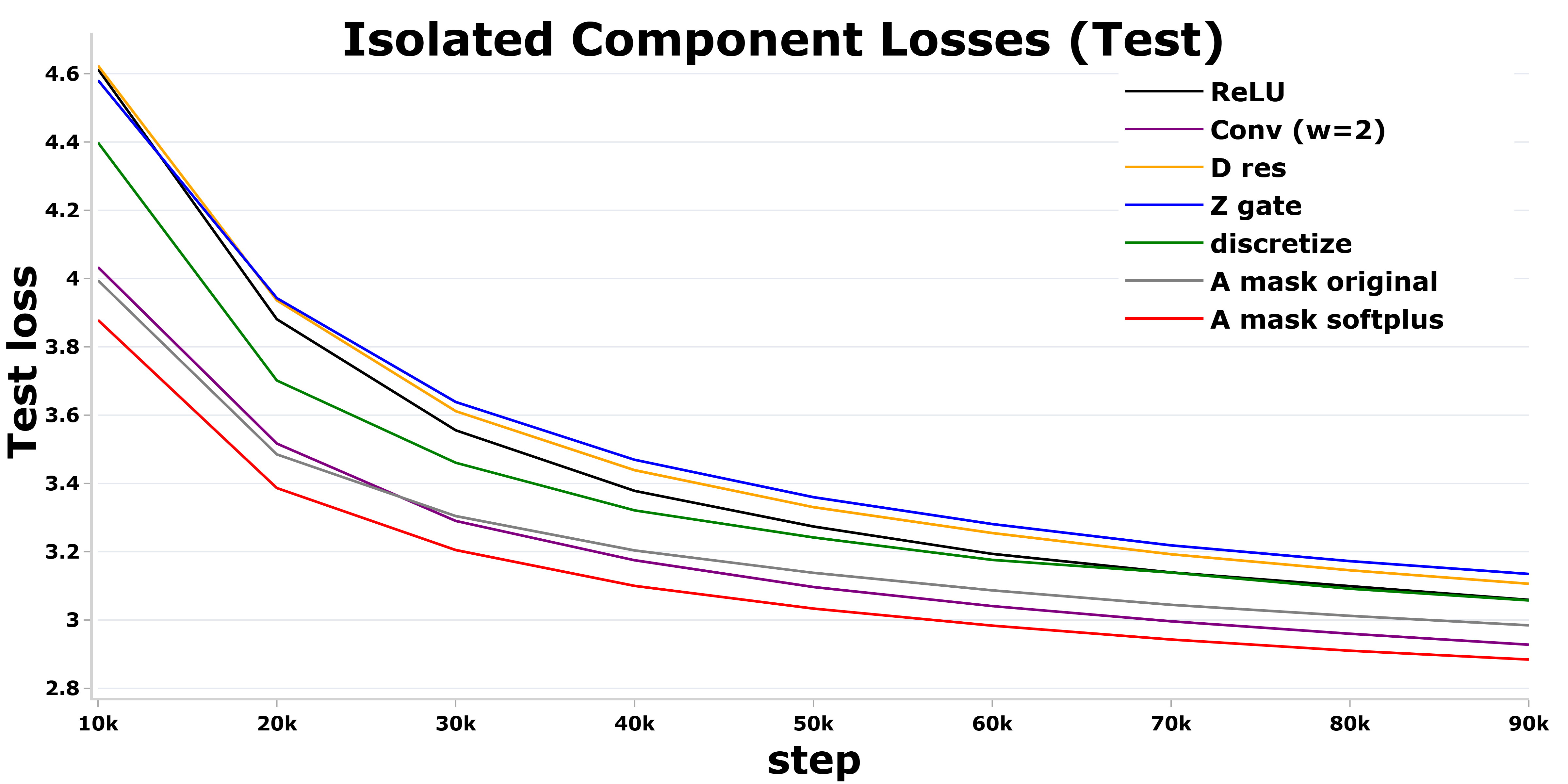}
        \caption{Isolated Mamba ablation}
        \label{fig:isolated}
    \end{figure}
\end{minipage}
\hfill
\begin{minipage}[c]{0.52\textwidth}
    \begin{table}[H]
    \centering
    \begin{tabular}{l|cc}
      \toprule
      Model & Train loss & Test loss\\
      \midrule
      SM Norm + ReLU & 3.0 & 3.03 \\
      Out Norm & 3.19 & 3.13 \\
      Out Norm + ReLU & 3.12 & 3.06 \\
      Out Norm + Conv (w=2) & 2.87 & 2.93 \\
      Out Norm + Conv (w=3) & 2.94 & 2.91 \\
      Out Norm + Conv (w=2) + SiLU & 2.95 & 2.9 \\
      Out Norm + D res & 3.18 & 3.11 \\
      Out Norm + Z gate & 3.2 & 3.13 \\
      Out Norm + value  discretize & 3.06 & 3.06 \\
      Out Norm + $A$-mask original & 2.94 & 2.98 \\
      Out Norm + $A$-mask softplus & \textbf{2.85} & \textbf{2.88} \\
      \bottomrule
    \end{tabular}
    \vspace{0.25em}
    \caption{Table of isolated component additions added to linear attention.}
    \label{tab:isolated}
    \end{table}
\end{minipage}
%
\begin{figure}[htbp]
    \centering
    \begin{subfigure}[b]{0.48\textwidth}
        \centering
        \includegraphics[width=\linewidth]{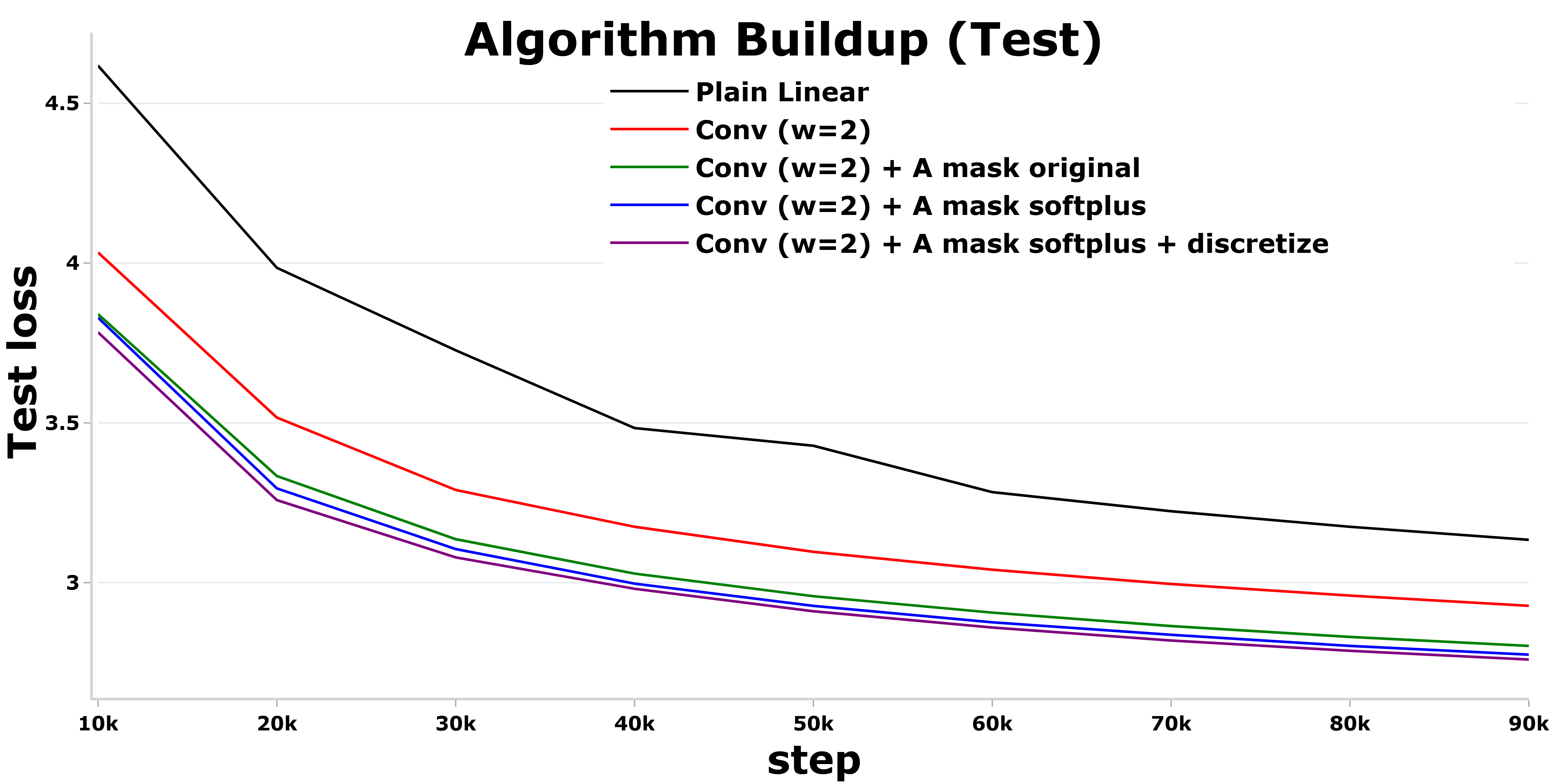}
        \caption{Main buildup}
        \label{fig:buildup_main}
    \end{subfigure}
    \hfill
    \begin{subfigure}[b]{0.48\textwidth}
        \centering
        \includegraphics[width=\linewidth]{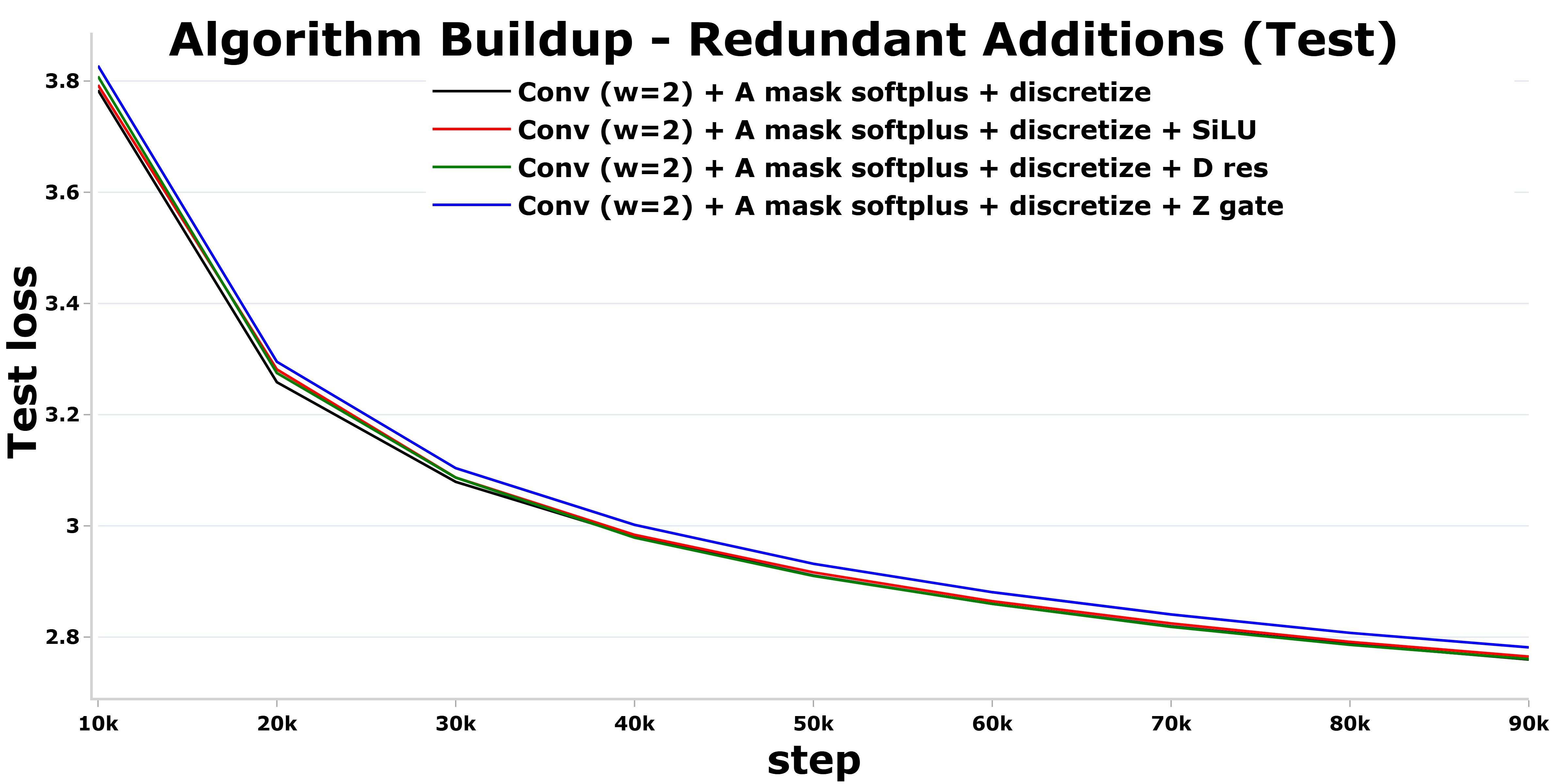}
        \caption{Redundant components after buildup}
        \label{fig:buildup_red}
    \end{subfigure}
    \caption{Investigations of major and minor build ups in constructing the simplified Mamba-2 architecture.}
\end{figure}
We ablate the rest of the components of Mamba-2 mentioned above in isolation. All isolated ablation test values can be found in Table \ref{tab:isolated}. Figure \ref{fig:isolated} also shows the results for ablating each component in isolation, graphically. The most impactful component was the $A$-mask, specifically the softplus variant. The original $A$-mask variant used in Mamba-2 is still quite impactful on model accuracy, but not as much as the softplus variant. The second most impactful component is the convolution. The rest of the isolated components result in minor accuracy gains compared to vanilla linear attention. We note that, of the collection of minor improvements, the value discretization method is most impactful. Even so, its impact is eclipsed by the $A$-mask and convolution components.

\section{Building Up to the Mamba-2S Base Model}

With each isolated component ablated, we want to buildup a minimal version of Mamba-2 that is as accurate as the original Mamba-2 algorithm (see Table \ref{tab:buildup}). We start with the \textit{output norm} variant of linear attention and build up the model on top of this base. While investigating all component interactions would be desirable, the number of combinations is too numerous. Instead, we build up the model using insights from the isolated component ablations. Figure \ref{fig:buildup_main} shows the main model buildup. To build up the new model, we start by combining the convolution and softplus $A$-mask as these two isolated components had the largest impact on accuracy, as shown in Figure \ref{fig:isolated}. We find the combination of these two components results in higher accuracy than the isolated components themselves and use this as the new base model. We then investigate adding a third component, finding that adding the time discretization parameter gives a small boost in accuracy. In Figure \ref{fig:buildup_red}, we show that adding other components do not help increase accuracy of the model. In fact, adding a Z gate slightly reduces accuracy of the model. All buildup investigations can be found in Table \ref{tab:buildup}.

\begin{table}[htbp]
\centering
\begin{tabular}{l|cc}
  \toprule
  Model & Train loss & Test loss\\
  \midrule
  Plain Linear & 3.19 & 3.13 \\
  Conv (w=2) & 2.87 & 2.93 \\
  Conv (w=2) + $A$-mask original & 2.81 & 2.8 \\
  Conv (w=2) + $A$-mask softplus & \textbf{2.67} & 2.77 \\
  \textit{Conv (w=2) + $A$-mask softplus + value discretize} & 2.7 & \textbf{2.76} \\
  Conv (w=2) + $A$-mask softplus + value discretize + SiLU & 2.73 & 2.76 \\
  Conv (w=2) + $A$-mask softplus + value discretize + D res & 2.79 & 2.76 \\
  Conv (w=2) + $A$-mask softplus + value discretize + Z gate & 2.72 & 2.78 \\
  \bottomrule
\end{tabular}
\vspace{0.25em}
\caption{Table of all additions (buildups) leading up to our final algorithm. Italics convey the components in \MambaSimple{}. }
\label{tab:buildup}
\end{table}

To keep the model both simple and accurate, our resulting method uses the softplus $A$-mask, an input convolution of size 2, and time discretization. This simplified Mamba-2 algorithm, which we call \MambaSimple{}, only retains necessary components as shown in Algorithm \ref{alg:mamba2_simple}.
\begin{algorithm}
\begin{algorithmic}
\caption{Proposed Mamba-2 Simplified Forward (\MambaSimple{}): no setup besides weights}
\label{alg:mamba2_simple}
\Require $h \in \mathbb{R}^{N, d}, W_{QKV} \in \mathbb{R}^{d, 3 \cdot (H \cdot d_h)}, W_{dt} \in \mathbb{R}^{d, H}, \textcolor{blue}{W_A \in \mathbb{R}^{d, H}}, W_{out} \in \mathbb{R}^{H \cdot d_h, d}$
\State \begin{align*}
Q, K, V &= \text{conv\_1d}(h \cdot W_{QKV}) &&\in \mathbb{R}^{H, N, d_h} \\[-4pt]
dt &= \text{softplus}(h \cdot W_{dt}) &&\in \mathbb{R}^{H, N} \\[-4pt]
A &=\textcolor{blue}{ -\text{softplus}(h \cdot W_A)} &&\in \mathbb{R}^{H, N} \\[-4pt]
V_{dt} &= V \odot dt && \in \mathbb{R}^{H, N, d_h} \\[-4pt]
A^{CS} &= \text{cumsum}( \textcolor{blue}{A}) &&\in \mathbb{R}^{H, N} \\[-4pt]
A^{M} &= \exp(A^{CS} - (A^{CS})^T) \quad \text{where } A^{M}_{ij} = \exp(A^{CS}_i - A^{CS}_j) &&\in \mathbb{R}^{H, N, N} \\
M_{ij} &= \begin{cases} 1, & \text{if } i \ge j\\ 0, & \text{if } i < j \end{cases} && M \in \mathbb{R}^{H, N, N} \\
y &= (Q K^T \odot A^M \odot M) \cdot V_{dt} &&\in \mathbb{R}^{H, N, d_h} \\[-4pt]
y' &= \text{vec}(y)_{(H, N, d_h) \longrightarrow (N, H \cdot d_h)} &&\in \mathbb{R}^{N, H \cdot d_h} \\[-4pt]
y_N &= \text{RMSNorm}(\textcolor{blue}{y'}) &&\in \mathbb{R}^{N, H \cdot d_h} \\[-4pt]
out &= y_N \cdot W_{out} &&\in \mathbb{R}^{N, d} \\[-4pt]
\end{align*}
\end{algorithmic}
\end{algorithm}
\section{Mamba-2 with a Squared Hidden State}

\begin{figure}[htbp]
    \centering
    \begin{subfigure}[b]{0.48\textwidth}
        \centering
        \includegraphics[width=\linewidth]{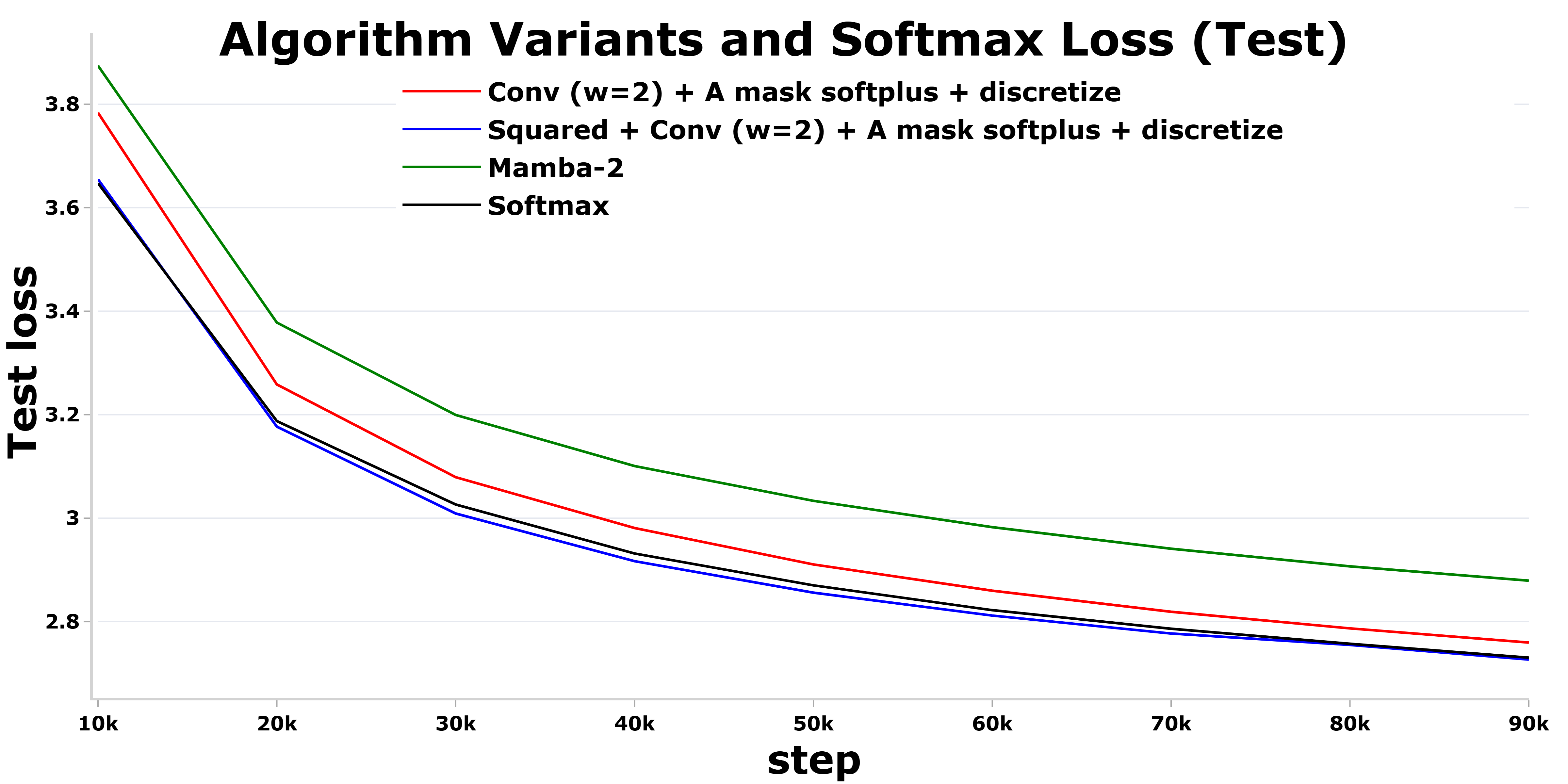}
        \caption{Small model performance comparison.}
        \label{fig:sm_comp}
    \end{subfigure}
    \hfill
    \begin{subfigure}[b]{0.48\textwidth}
        \centering
        \includegraphics[width=\linewidth]{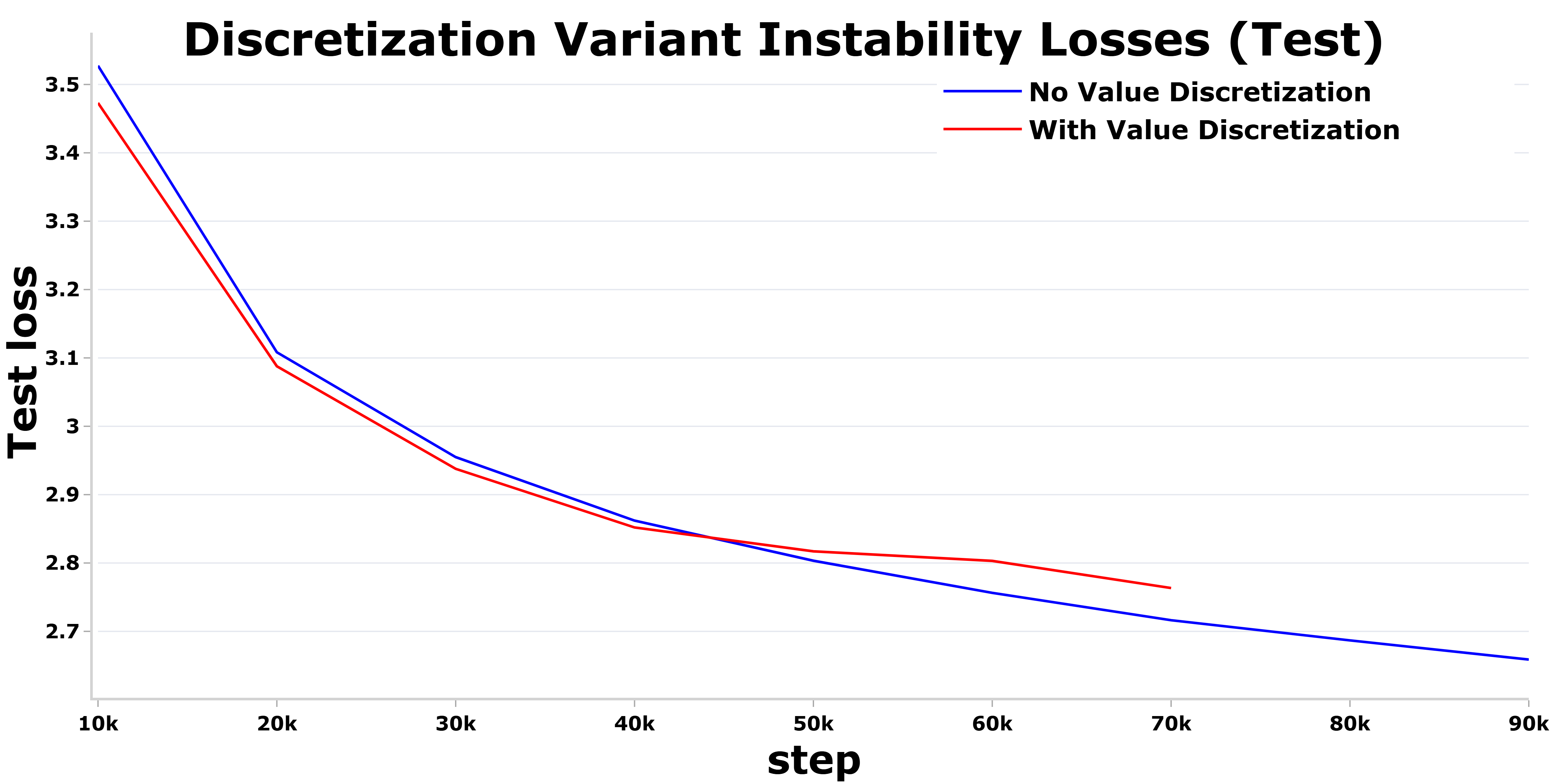}
        \caption{Instability in the discretization variant.}
        \label{fig:unstable}
    \end{subfigure}
    \caption{Experimental results comparing accuracy and training stability.}
\end{figure}

\begin{figure}[htbp]
    \centering
    \begin{subfigure}[b]{0.32\textwidth}
        \centering
        \includegraphics[width=\linewidth]{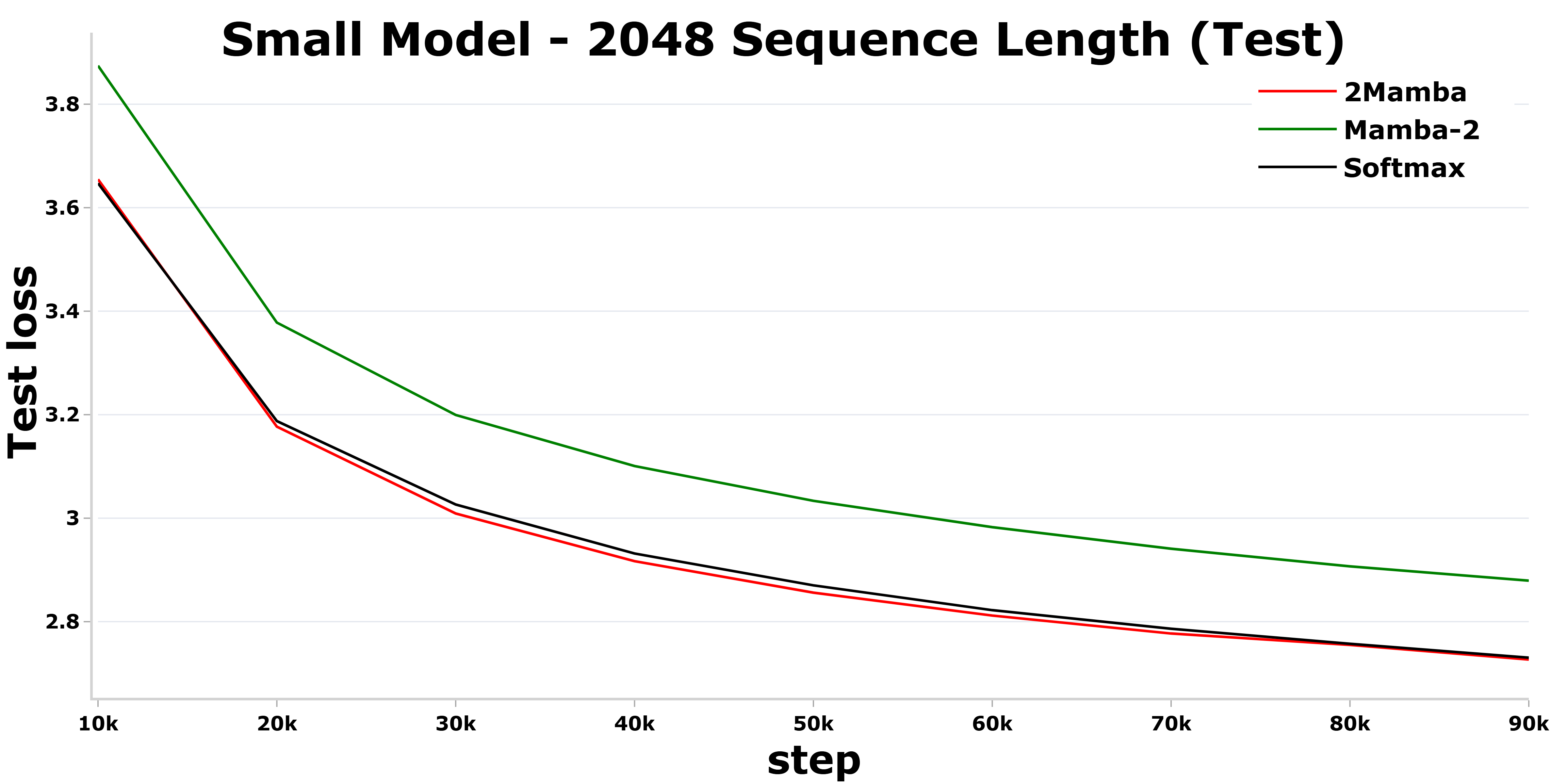}
    \end{subfigure}
    \hfill
    \begin{subfigure}[b]{0.32\textwidth}
        \centering
        \includegraphics[width=\linewidth]{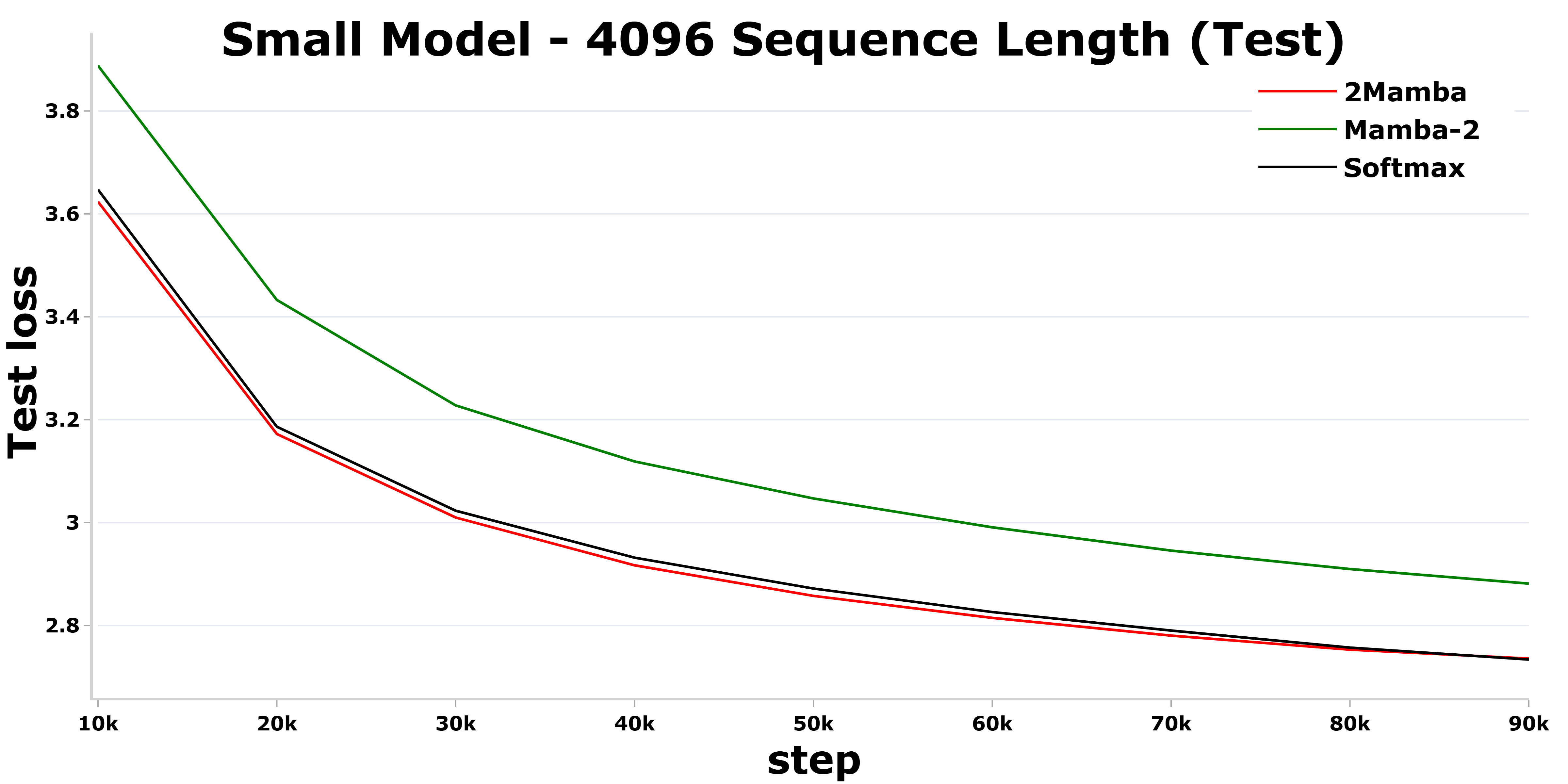}
    \end{subfigure}
    \hfill
    \begin{subfigure}[b]{0.32\textwidth}
        \centering
        \includegraphics[width=\linewidth]{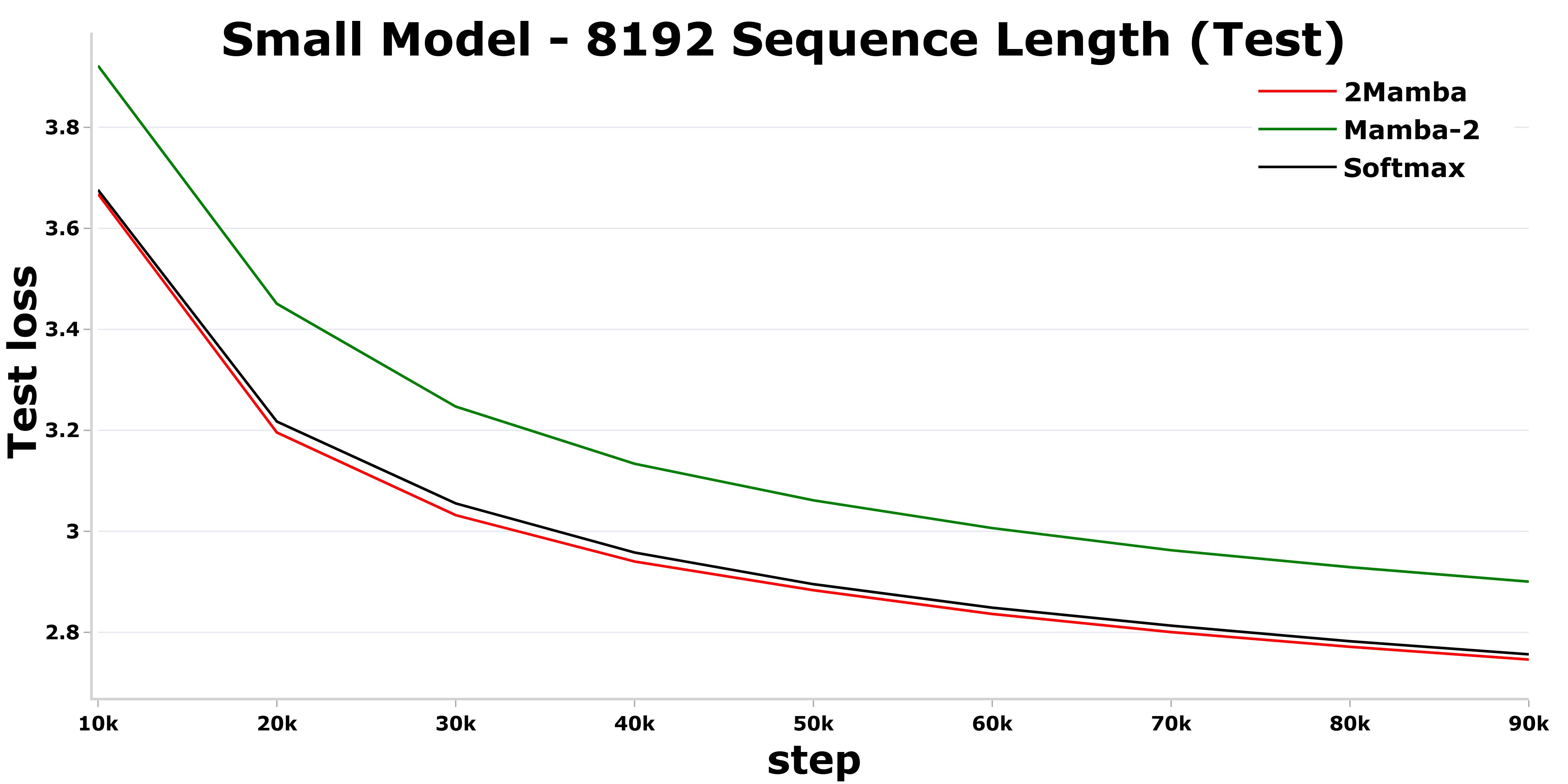}
    \end{subfigure}

    \vspace{1em} 

    \begin{subfigure}[b]{0.32\textwidth}
        \centering
        \includegraphics[width=\linewidth]{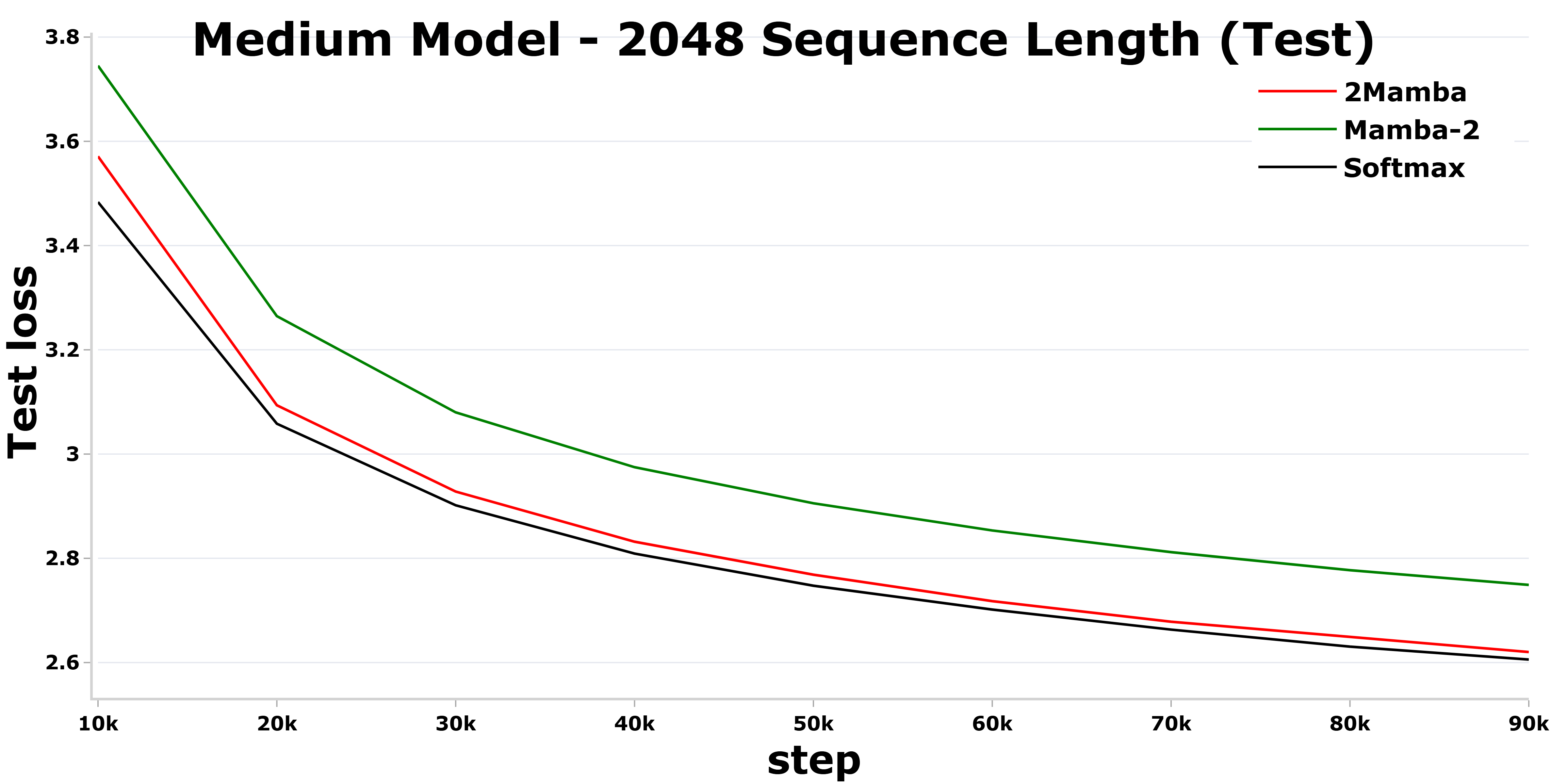}
    \end{subfigure}
    \hfill
    \begin{subfigure}[b]{0.32\textwidth}
        \centering
        \includegraphics[width=\linewidth]{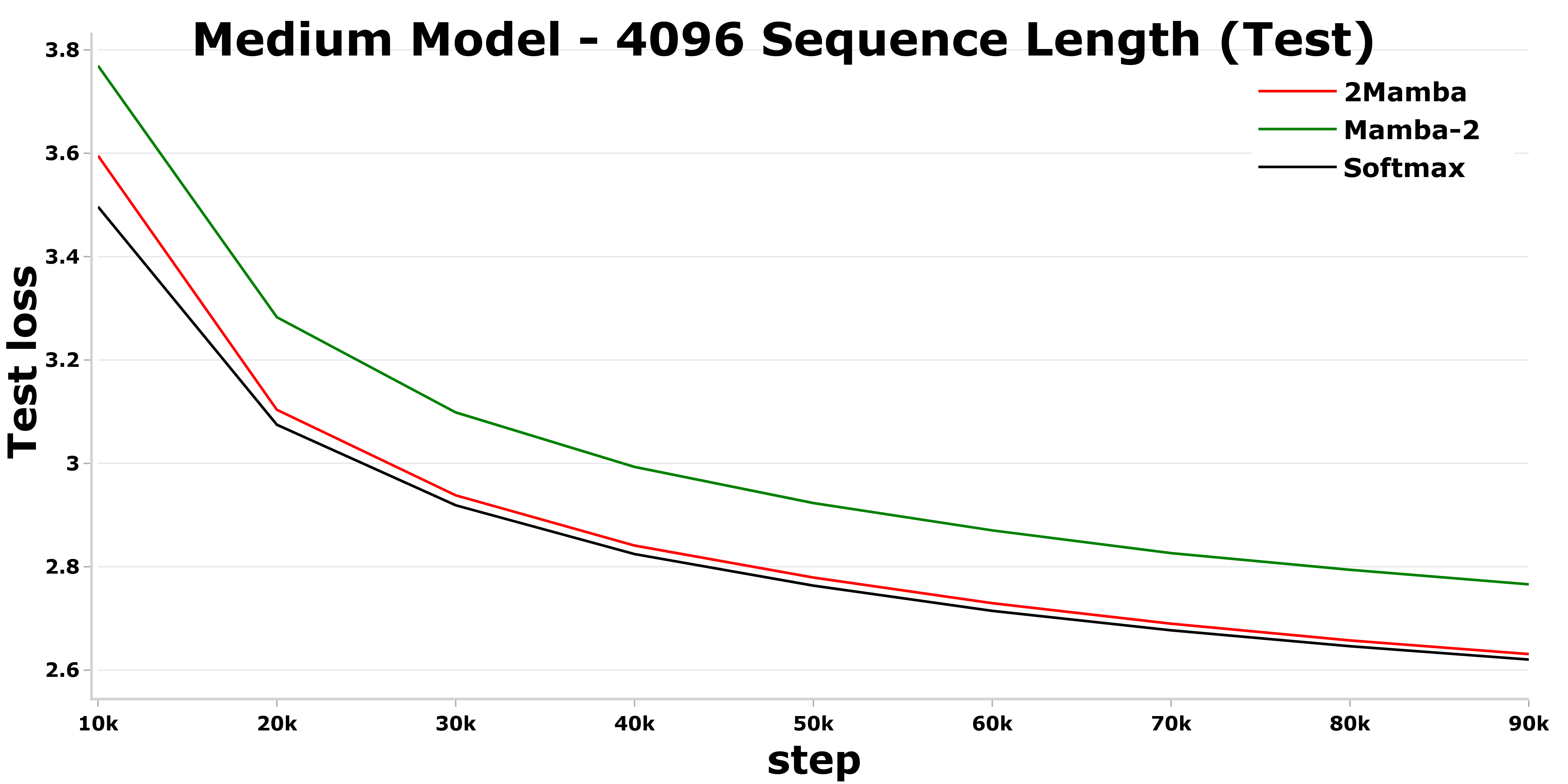}
    \end{subfigure}
    \hfill
    \begin{subfigure}[b]{0.32\textwidth}
        \centering
        \includegraphics[width=\linewidth]{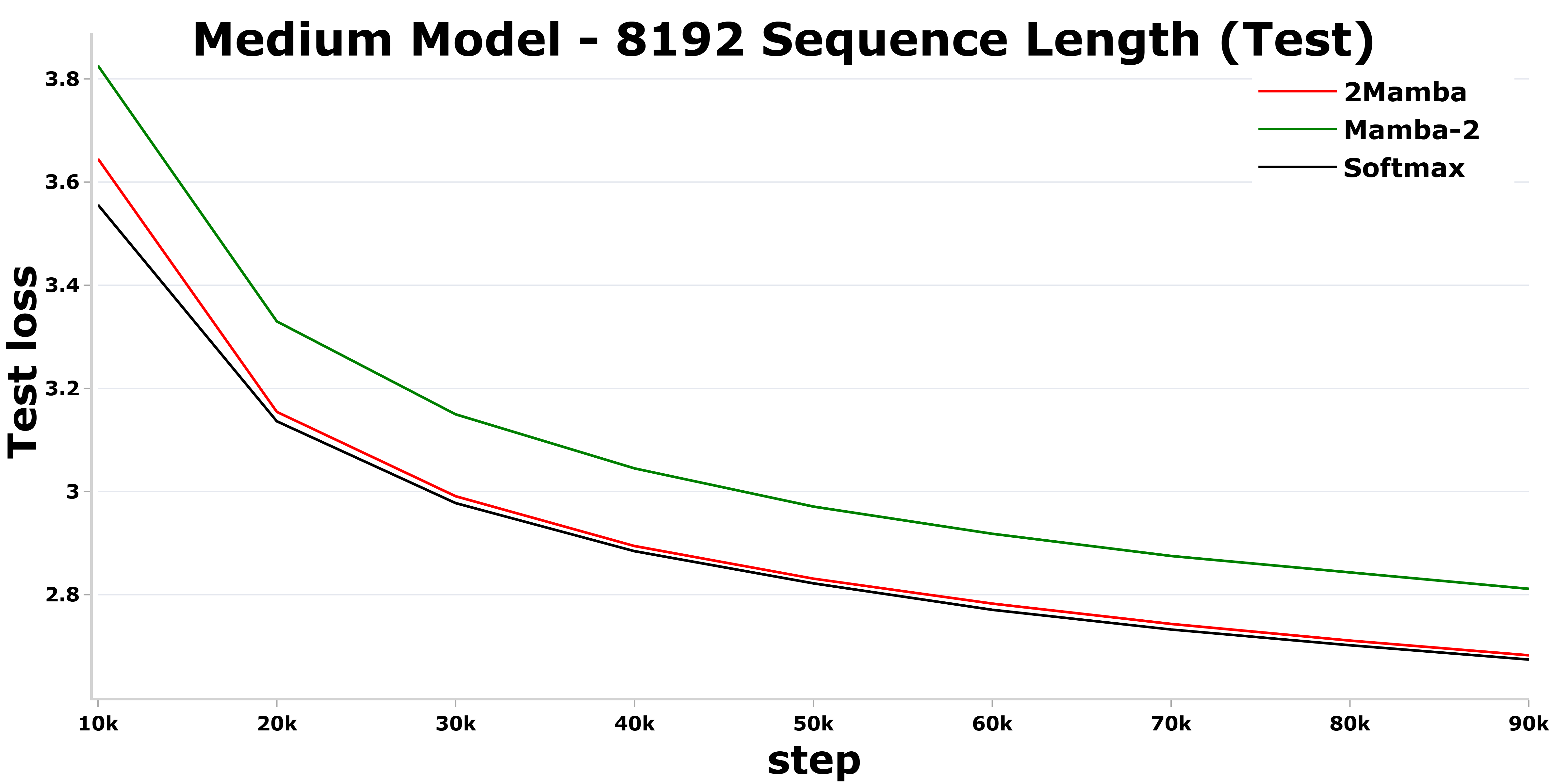}
    \end{subfigure}
    
    \caption{\textbf{First row}: Small model ($\sim$300M params) test loss for Mamba-2, \AlgoName{}, and softmax attention trained on 2048, 4096, and 8192 sequence lengths. \textbf{Second row}: Medium model ($\sim$700M params) test loss for Mamba-2, our proposed algorithm, and softmax attention trained on 2048, 4096, and 8192 sequence lengths}
    \label{fig:full_grid_comparison}
\end{figure}

While \MambaSimple{} is quite strong in accuracy as seen in Figure \ref{fig:sm_comp}, it still falls short of softmax attention. (\citet{on_the_expr_of_sm_attn}) showed that squaring the $QK$ inner product results in a significant accuracy boost. Additionally, squaring the $QK$ inner product results in a positive inner product space image. This space can adopt an online normalization algorithm like \textit{softmax normalization}, which is more stable than \textit{output normalization}. Using the simplified Algorithm \ref{alg:mamba2_simple}, we find that squaring the $QK$ inner product does result in a significant accuracy gain. As seen in Figure \ref{fig:sm_comp}, squaring the inner product results in softmax level accuracy while being more memory efficient than softmax attention for sufficient sequence length. Specifically, from \MambaSimple{} (Algo \ref{alg:mamba2_simple}), we remove the RMSNorm, add softmax normalization, and replace the $QK$ inner product with a squared inner product. We opt to use softmax normalization over output normalization as softmax normalization is more stable when using the online softmax algorithm (\citet{online_sm}). Online softmax normalization, like Flash Attention (\citet{flash_attn}), requires an exponential to update the max statistic over blocks. As the $A$-mask has an exponential, online softmax-like normalization can still be effective when the max statistic is stored in the $A$-mask as opposed to the $e^{QK}$ inner product, as used in online softmax. The gradients for the forward and backward pass are supplied in Appendix \ref{app:gradients}. Full Triton kernels can be found in our codebase, which are necessary for training.
\\
The time discretization operation does improve the model, however the accuracy gains are minimal compared to adding the convolution and softplus $A$-mask. For smaller models, training is stable. For the medium model, we found that the magnitude of the values significantly increased across update steps, leading to numerical instability and divergence in training as seen in Figure \ref{fig:unstable}. Training a larger model with discretization can be somewhat stabilized---however this requires forcing triton to use IEEE FP32 input precision which is about 8x slower (\citet{tf32}) than the TF32 precision (a reduced precision mantissa compared to the standard IEEE FP32). A reasonable middle ground is a mix of input precision of FP32x3 for dot operations that need more precision and TF32 for others, which is still slower than full TF32 precision but is slightly more stable. Training in pure FP32 PyTorch also results in stable training, but is considerably slower. While manipulating the kernel can result in more stable training, to keep the squared variant numerically stable for low precision kernels, we remove the time discretization operation for the medium model. 
\vspace{0.5em}
\\
Our proposed squared algorithm, which we name \AlgoName{}, can be found in Algorithm \ref{alg:squared_mamba}. As seen in Figure \ref{fig:full_grid_comparison}, across various model sizes and sequence lengths, \AlgoName{} is nearly as accurate as softmax attention. One aspect worth further discussion is the placement of the square operation. We specifically place it on the $QK$ inner product to increase the inner product dimension. An alternative placement could also include the $A$-mask. However, we find that including the $A$-mask does not result in any accuracy gains. While squaring after the $A$-mask application does result in a squared inner product on $Q$ and $K$, it also results in squared terms in the $A$-mask, which is unnecessary. Intuitively, the $A$-mask should be applied directly to attention scores, which come from the squared inner product of $Q$ and $K$.
\begin{algorithm}
\begin{algorithmic}
\caption{\AlgoName{} Algorithm}
\label{alg:squared_mamba}
\Require $h \in \mathbb{R}^{N, d}, W_{QKV} \in \mathbb{R}^{d, 3 \cdot (H \cdot d_h)}, W_A \in \mathbb{R}^{d, H}, W_{out} \in \mathbb{R}^{H \cdot d_h, d}$
\State \begin{align*}
Q, K, V &= \text{conv\_1d}(h \cdot W_{QKV}) &&\in \mathbb{R}^{H, N, d_h} \\[-4pt]
A &= -\text{softplus}(h \cdot W_A) &&\in \mathbb{R}^{H, N} \\[-4pt]
A^{CS} &= \text{cumsum}( A) &&\in \mathbb{R}^{H, N} \\[-4pt]
A^{M} &= \exp(A^{CS} - (A^{CS})^T) \quad \text{where} \quad A^{M}_{ij} = \exp(A^{CS}_i - A^{CS}_j) && \in \mathbb{R}^{H, N, N}, \\[-4pt]
M_{ij} &= \begin{cases} 1, & \text{if } i \ge j\\ 0, & \text{if } i < j \end{cases} && M \in \mathbb{R}^{H, N, N}\\[-0pt]
y &= ((Q K^T)^{\textcolor{blue}{2}} \odot A^M \odot M) \cdot V &&\in \mathbb{R}^{H, N, d_h} \\[-0pt]
\textcolor{blue}{N} &\textcolor{blue}{= \sum_j ((Q K^T)^2 \odot A^M \odot M)} &&\textcolor{blue}{\hspace{0.25em}\in \mathbb{R}^{H, N}} \\[-0pt]
\textcolor{blue}{y_N} &\textcolor{blue}{= y / N} &&\textcolor{blue}{\hspace{0.25em}\in \mathbb{R}^{H, N, d_h}} \\[-4pt]
y'_N &=\text{vec}(y_N)_{(H, N, d_h) \longrightarrow (N, H \cdot d_h)} &&\in \mathbb{R}^{N, H \cdot d_h} \\[-4pt]
out &= y'_N \cdot W_{out} &&\in \mathbb{R}^{N, d} \\[-4pt]
\end{align*}
\end{algorithmic}
\end{algorithm}
\subsection{\AlgoName{} Algorithm Efficiency}
\label{sec:rnn_comp_red}
\begin{figure}[htbp]
    \centering
    \includegraphics[width=0.6\textwidth]{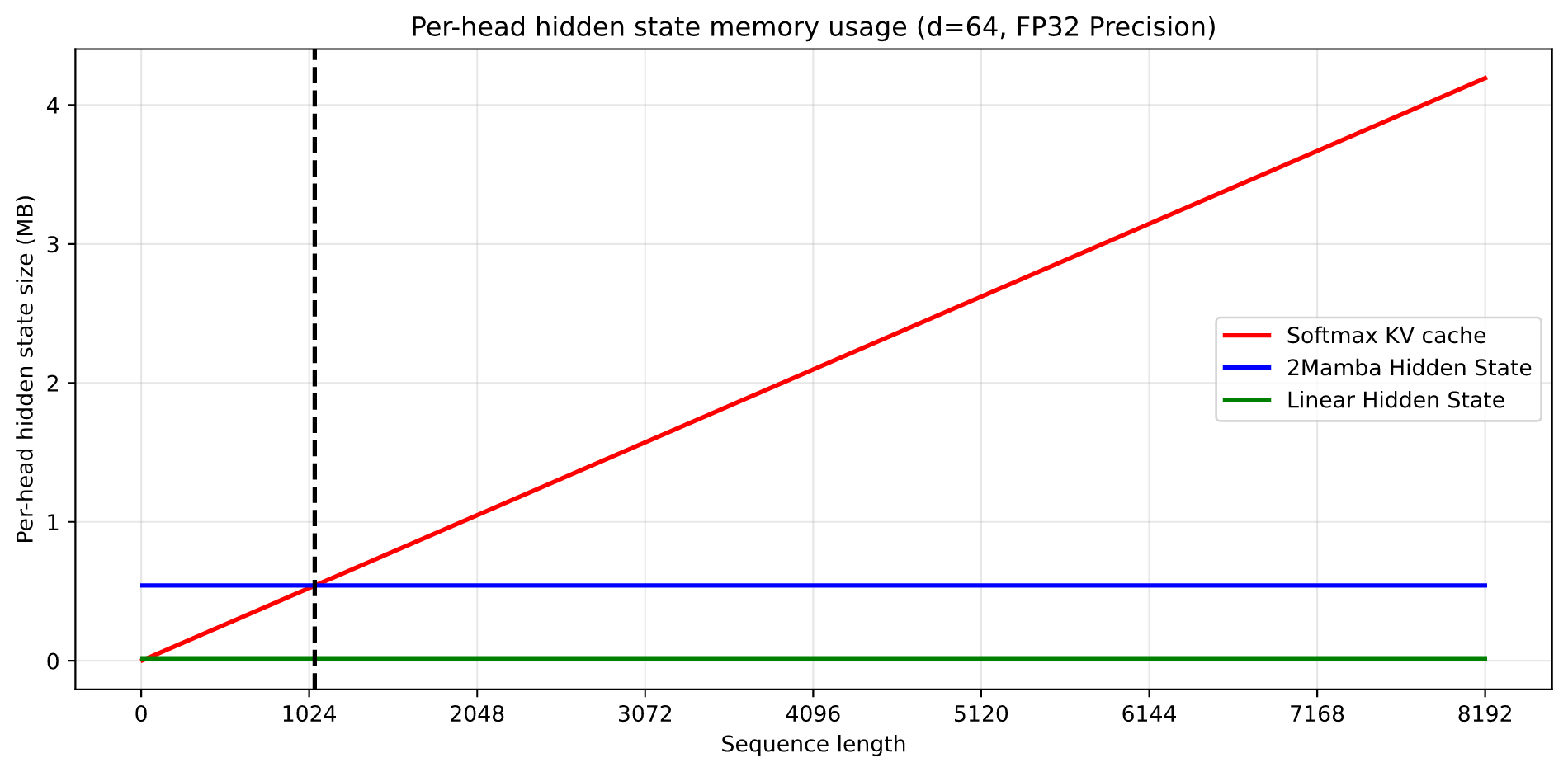}
    \caption{Memory usage of a single head of softmax attention, \AlgoName{}, and linear attention, utilizing softmax-like normalization and a convolution with kernel size 2.}
    \label{fig:mem_usage}
\end{figure}

While squaring the inner product does result in a large accuracy gain, we note that the recurrent state requires approximately the square of the space as the original linear variant. When squaring the inner product or taking the Kronecker product of $Q$ and $K$, and then performing the inner product, some terms are repeated due to the commutativity of the multiplication operation in the inner product. This means that as opposed to doing a Kronecker product on the queries and keys, one can do an operation that calculates all unique second-order product terms and calculate their multiplicity via the multinomial theorem. We provide a Triton kernel for creating a vector of second order products of a given vector.\footnote{\href{https://github.com/gmongaras/Triton-Efficient-Kronecker-Product}{https://github.com/gmongaras/Triton-Efficient-Kronecker-Product}} The number of terms, as seen in equation \eqref{eq:multinomial} is equal to $\frac{d(d+1)}{2}$, which is less than half the elements within the vector obtained from naively calculating all $d^2$ elements.\footnote{For notational brevity, we denote the head dimension as $d$ in this section.}
\begin{equation}
    \label{eq:multinomial}
    num\_terms = \genfrac(){0pt}{0}{2+d-1}{d-1} = \genfrac(){0pt}{0}{d+1}{d-1} = \frac{(d+1)!}{2(d-1)!} = \frac{(d+1)\cdot d \cdot (d-1)!}{2(d-1)!} = \frac{d\cdot (d+1)}{2}
\end{equation}
As the complexity of softmax attention is linear in memory due to the $KV$ cache, there exists a sequence length such that squaring the inner product requires less memory than softmax attention. A $KV$ cache requires memory on the order of $2Ld$ for a single attention head. By using equation \eqref{eq:multinomial} for the key dimension and multiplying by $d$ for the value dimension, a hidden state for squared linear attention requires memory on the order of $d^2(d+1)/2$ per attention head. Adding a convolution adds an extra $3d$ memory per attention head and using softmax normalization adds an extra $d(d+1)/2$ memory per attention head. In total, this results in a hidden state size of $\frac{d(d+1)^2}{2} + 3d$ per attention head. The inequality in equation \eqref{eq:squared_ineq} provides the necessary sequence length, $N$, such that any sequence length longer than $N$ is more memory efficient using a second-order hidden state than using softmax attention.
\begin{equation}
    \label{eq:squared_ineq}
    2Nd > \frac{d(d+1)^2}{2} + 3d \qquad \longrightarrow \qquad N > \frac{(d+1)^2}{4} + \frac{3}{2}
\end{equation}
In practice, we used a head dimension of 64. According to inequality in equation \eqref{eq:squared_ineq}, a sequence length greater than $\frac{(64+1)^2}{4} + \frac{3}{2} \approx 1058 $ would result in a lower memory usage than softmax attention. To ensure the squared model retains good accuracy beyond this sequence length, we examine models with a context size up to 8192 tokens. As such, squaring the $QK$ inner product is less memory constrained than softmax attention past the sequence length of $1058$ while matching softmax level accuracy. This memory result is verified by running both softmax attention and \AlgoName{} while accumulating the raw hidden state memory usage from each model, as seen in figure \ref{fig:mem_usage}. Our codebase has a state caching inference implementation for both algorithms. The inference algorithm is also supplied in Appendix \ref{app:inf}.
\subsection{Effective Context Usage}

To confirm \AlgoName{} uses its context effectively, we employ the The Needle in a Haystack (NIAH) test, which measures ability of the model to retrieve a specific, relevant fact (the ``needle'') inserted into a long prompt (the ``haystack''). That is, it assesses how well a model recalls information placed at different locations within large prompts. In our testing, we train a model for $400,000$ steps on a batch size of $64$ with a maximum context size of $8,192$ tokens. During this long training run, we plot the test loss, as seen in Figure \ref{fig:long_test_loss}, showing that \AlgoName{} is still competitive with softmax attention for long training runs. After training for $400,000$ steps, we evaluate our model on Nanotron's needle in a haystack test\footnote{\href{https://huggingface.co/datasets/nanotron/simple\_needle\_in\_a\_hay\_stack}{https://huggingface.co/datasets/nanotron/simple\_needle\_in\_a\_hay\_stack}} on context lengths from about $1,024$ characters to about $16,384$ characters. This test hides a password inside a large corpus of irrelevant text. The model must memorize the password, and repeat it after parsing the large corpus of text. This benchmark tests how well a model uses its context window and for how long it can store important information. As seen in Figure \ref{fig:niah}, \AlgoName{} method is slightly better than softmax in context retrieval, and much better than Mamba-2. This result provides evidence that \AlgoName{} extends beyond NTP tasks and can utilize its context effectively.

\noindent
\begin{minipage}[c]{0.45\textwidth}
    \begin{figure}[H]
        \centering
        \includegraphics[width=\linewidth]{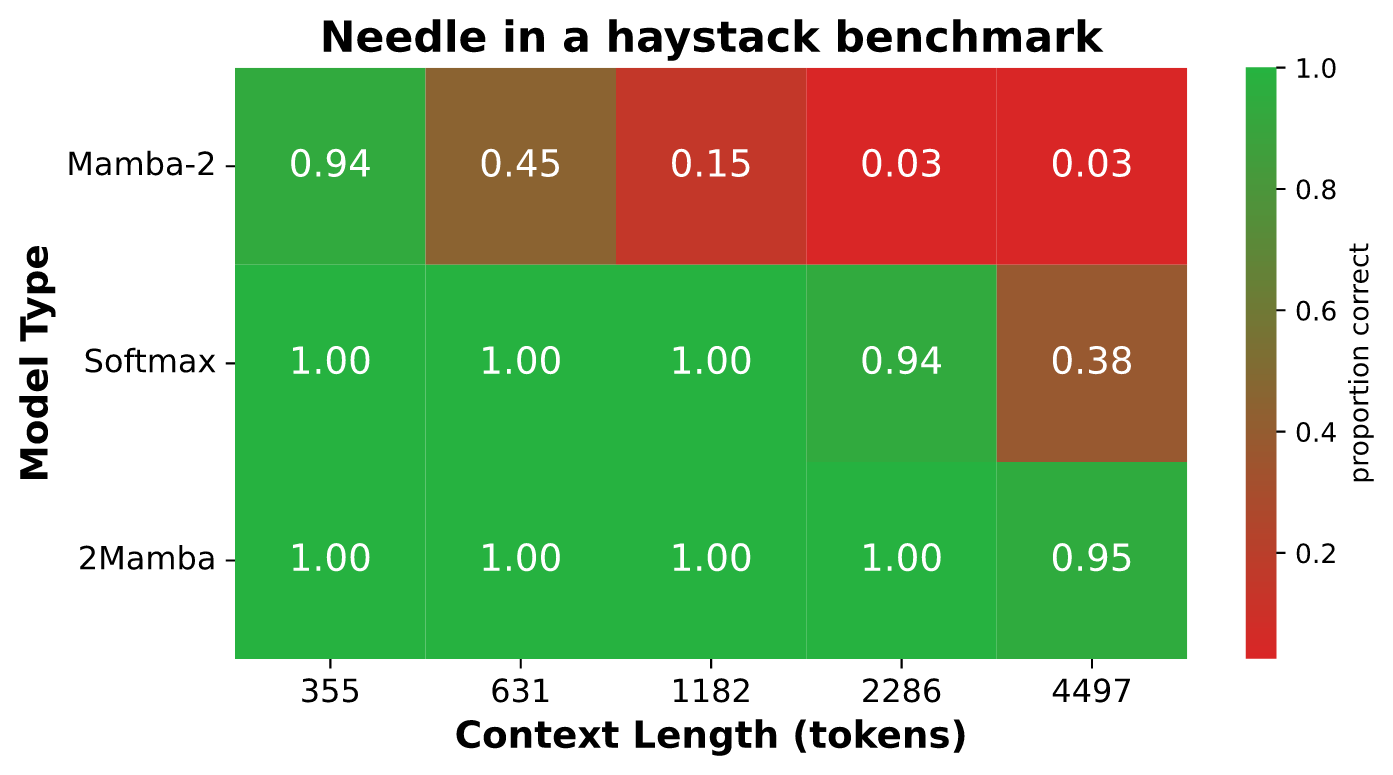}
        \caption{One-shot NIAH benchmark results using maximum likelihood sampling over 1000 sequences. Green indicates the predicted needle was correct more often while red indicates the predicted needle was predicted less often. Scores are indicative of the proportion of sequences in which the model predicted the needle correctly.}
        \label{fig:niah}
    \end{figure}
\end{minipage}
\hfill
\begin{minipage}[c]{0.45\textwidth}
    \begin{figure}[H]
        \centering
        \includegraphics[width=\linewidth]{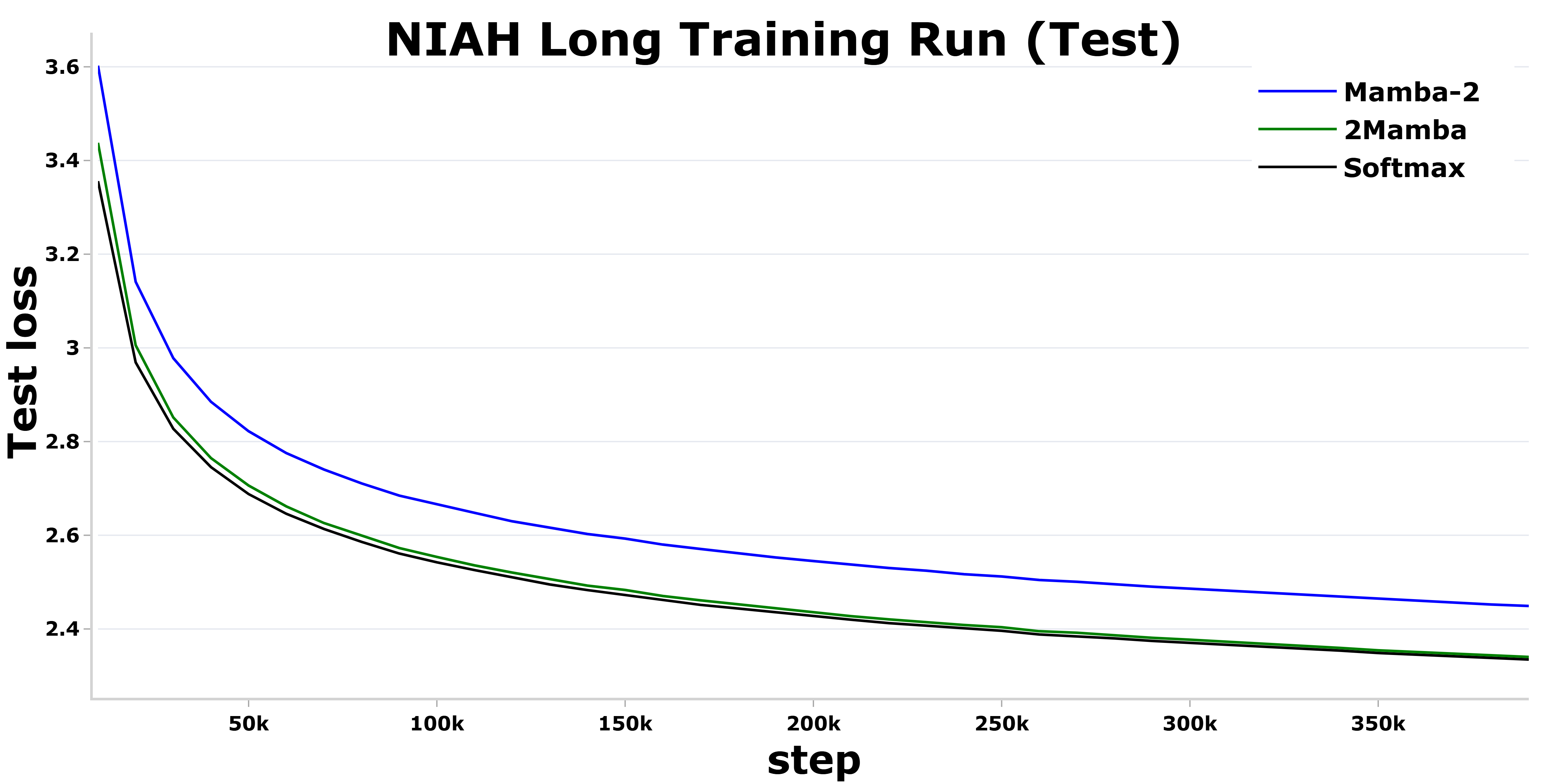}
        \caption{Long training run of various models with a batch size of 64 on $8,192$ max sequence length.}
        \label{fig:long_test_loss}
    \end{figure}
\end{minipage}
\section{\AlgoName{} With an Exponentiated Hidden State}
The property of squaring the inner product was used to make Mamba-2 more accurate while remaining efficient. As shown in (\citet{on_the_expr_of_sm_attn}), adding higher orders typically results in additional accuracy gains (though diminishing with each new term). In the limiting case, one can use the exponential function which results in softmax attention. Therefore, we ask: If the exponential function on normal linear attention gives softmax attention, then what would happen if the $QK$ inner product of \AlgoName{} was exponentiated as opposed to squared?
\begin{figure}[htbp]
    \centering
    \begin{subfigure}[b]{0.32\textwidth}
        \centering
        \includegraphics[width=\linewidth]{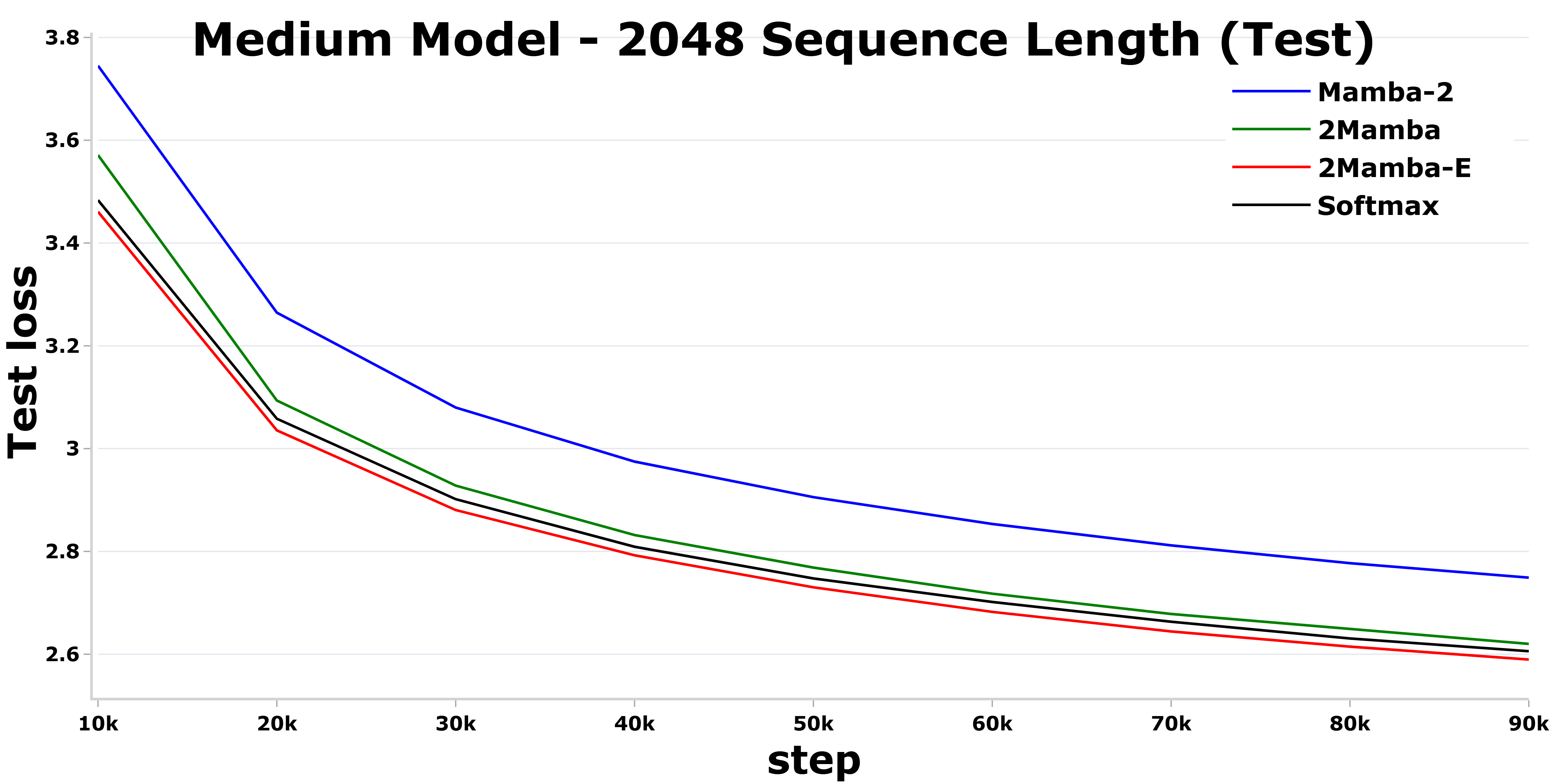}
    \end{subfigure}
    \hfill
    \begin{subfigure}[b]{0.32\textwidth}
        \centering
        \includegraphics[width=\linewidth]{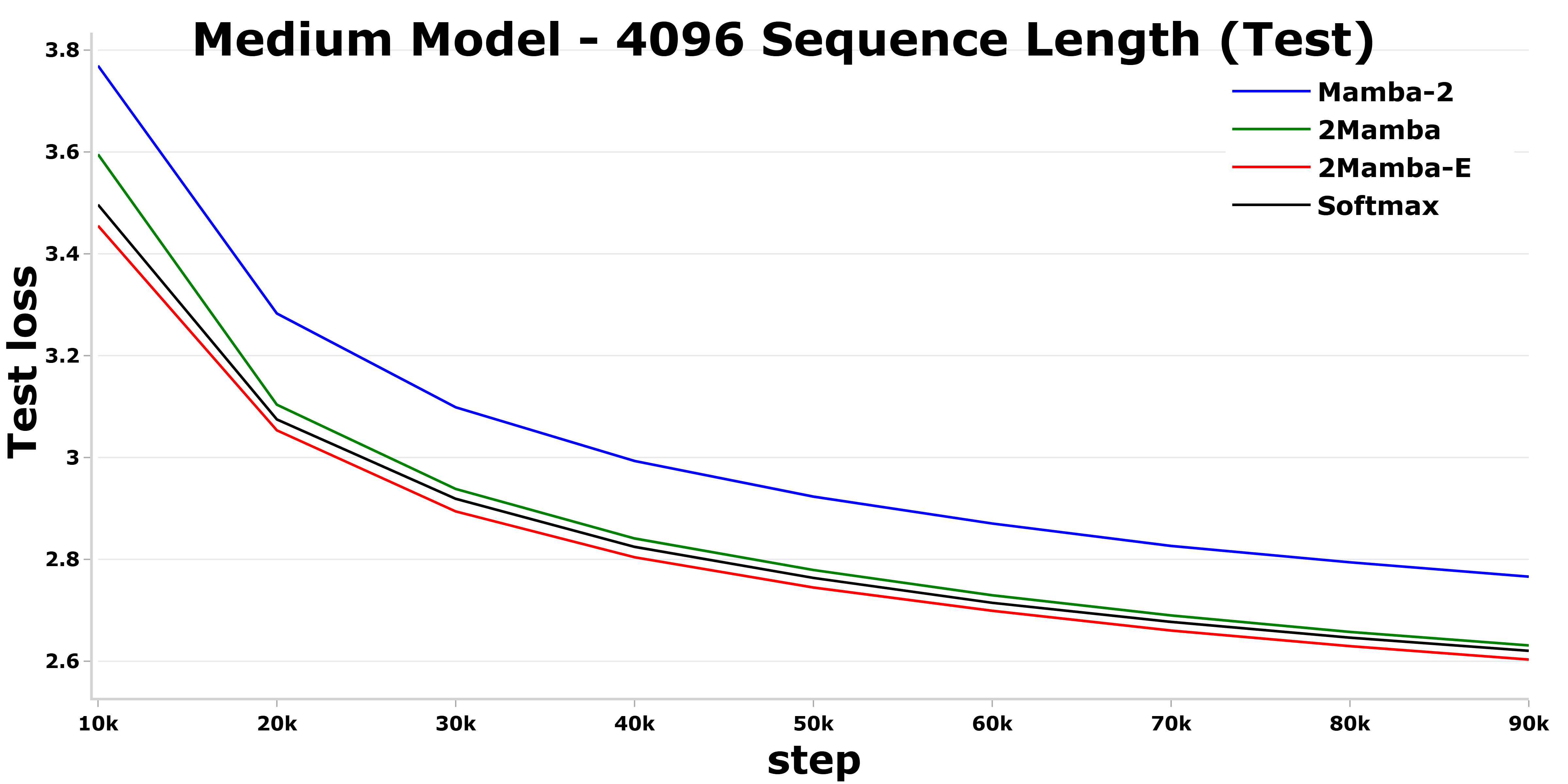}
    \end{subfigure}
    \hfill
    \begin{subfigure}[b]{0.32\textwidth}
        \centering
        \includegraphics[width=\linewidth]{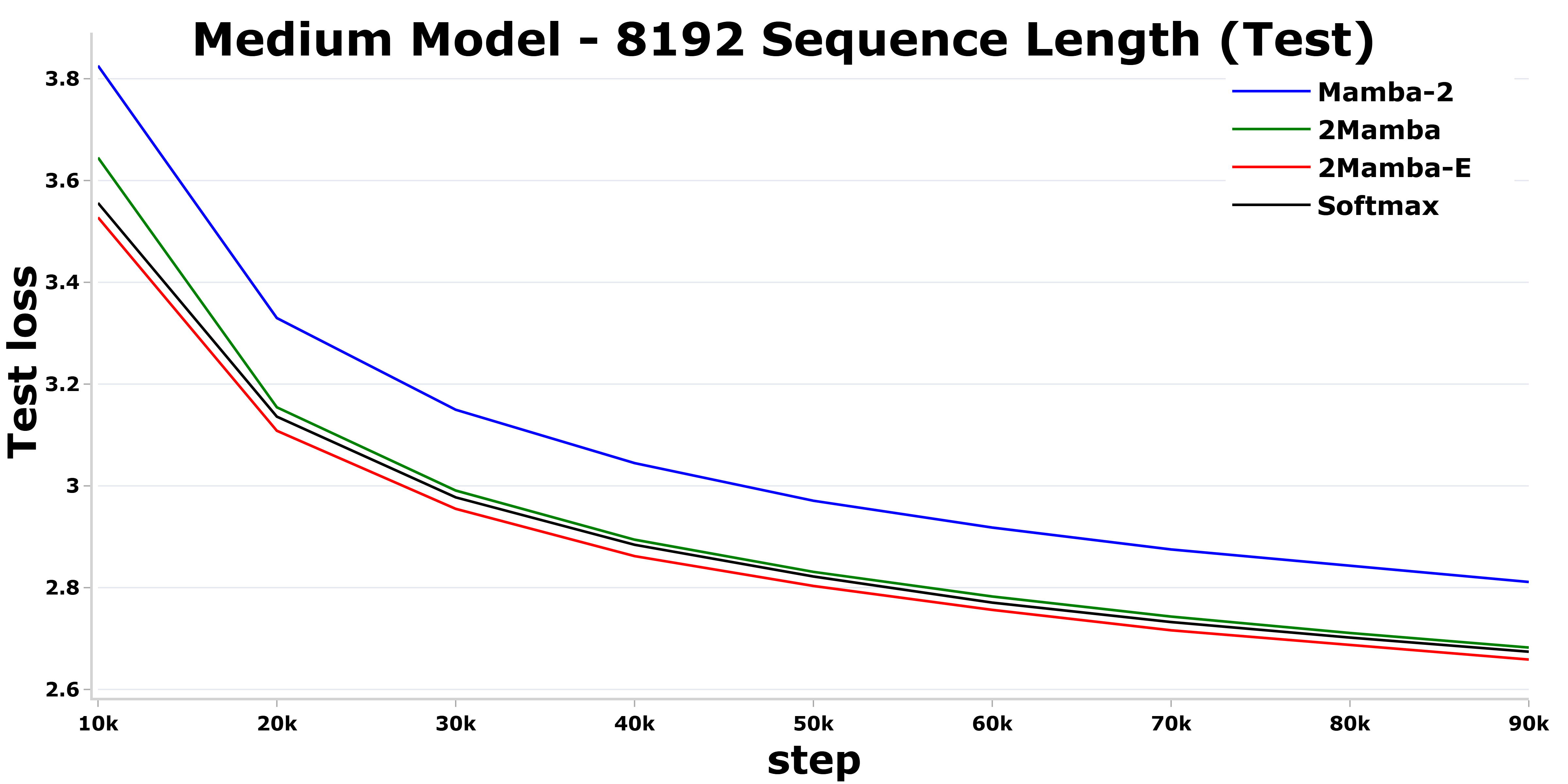}
    \end{subfigure}
    
    \caption{Medium model ($\sim$700M params) test loss for Mamba-2, \AlgoNameExp{}, and softmax attention trained on 2048, 4096, and 8192 sequence lengths.}
    \label{fig:exp}
\end{figure}
As seen in Figure \ref{fig:exp}, exponentiating the inner product of $Q$ and $K$ results in slightly better accuracy than softmax attention. We name the exponentiated form \AlgoNameExp{}. One can interpret this algorithm as either exponentiating \AlgoName{} or as softmax attention with the softplus $A$-mask and input convolution. The latter explanation seems more intuitive and satisfying when considering traditional transformer literature. Adding an $A$-mask, or decay mask, to softmax attention has also been proposed in the Forgetting Transformer (\citet{fox_forgetting_trans}). The main difference between the forgetting transformer and the algorithm we use is the softplus $A$-mask construction. The forgetting transformer uses $\log(\text{sigmoid}(x))$ as the function for the softplus $A$-mask values while we use $-\text{softplus}(-x)$. It turns out these are nearly equivalent (only differing by sign) as seen in equation \ref{eq:fox_equal}.
\begin{equation}
    \label{eq:fox_equal}
    \log(\text{sigmoid}(x)) = \log\left(\frac{1}{1+e^{-x}}\right) = -\log(1+e^{-x}) = -\text{softplus}(-x)
\end{equation}
From this perspective, our work on exponentiated hidden states also provides an intuitive connection between SSM models, like Mamba-2, and transformer variants like the forgetting transformer. 
\section{Conclusion and Future Work}
%
%
This work examines the importance of each individual component of Mamba-2 to build up a minimal implementation, \MambaSimple{}. From this base model, we take the second order hidden state to achieve a model, \AlgoName{}, that is as accurate as softmax attention, yet constant in memory. Additionally, we show that our implementation, when exponentiated, results in an architecture, \AlgoNameExp{}, that is better than softmax attention and show that this implementation is similar to the forgetting transformer (\citet{fox_forgetting_trans}). \AlgoName{} is a step forward in bridging the gap between linear attention architectures and softmax attention, while being more efficient in practice.
\vspace{0.5em}
\\
The exploration of this work closely examined the architecture choices made in Mamba-2 (\citet{mamba2}). Another parallel work that improves linear attention is DeltaNet (\citet{deltanet}). Mamba-2 and DeltaNet have been combined in Gated DeltaNet (\citet{gated_deltanet}). Similarly, future work can look into adding DeltaNet to \AlgoName{} to further improve the algorithm.
\vspace{0.5em}
\\
Future work can also look into the hidden state created by squaring the inner product of the queries and keys. While we stick with a static hidden state size, future work can examine how varying the hidden state size affects downstream accuracy and if squaring the inner product space results in better or similar accuracy than using a query and key projection to the same size. This future work can make \AlgoName{} more efficient in terms of memory by optimizing the hidden state size.

\bibliographystyle{plainnat}
\bibliography{references}

\appendix
\section{Inference Algorithm}
\label{app:inf}
Inference requires keeping around a few hidden states. $H^{\uparrow}_{t-1}$ is the numerator, keeping all key-value products. $H^{\downarrow}_{t-1}$ is the denominator, only keeping the keys and is only necessary for softmax-like normalization. With an input convolution, one must keep around $window\_size-1$ number of queries, keys, and values. The code below shows a window size of 2, so only one query, key, and value must be cached.
\begin{algorithm}
\begin{algorithmic}
\caption{\AlgoName{} Inference}
\label{alg:inference}
\Require $h \in \mathbb{R}^{d}, W_{QKV} \in \mathbb{R}^{d, 3 \cdot (H \cdot d_h)} W_A \in \mathbb{R}^{d, H}, W_{out} \in \mathbb{R}^{H \cdot d_h, d}$
\Require $H^{\uparrow}_{t-1} \in \mathbb{R}^{H, D, d_h}, H^{\downarrow}_{t-1} \in \mathbb{R}^{H, D}, q_{t-1} \in \mathbb{R}^{H, d_h}, k_{t-1} \in \mathbb{R}^{H, d_h}, v_{t-1} \in \mathbb{R}^{H, d_h}$
\Require $D = \frac{d_h (d_h + 1)}{2} \le d_h^2$
\State \begin{align*}
q_t, k_t, v_t &= h_t W_{QKV} &&\in \mathbb{R}^{H, d_h} \\[-4pt]
q^c_t, k^c_t, v^c_t &= \text{conv\_1d}(q_t, k_t, v_t, q_{t-1}, k_{t-1}, v_{t-1}, \text{window\_size}=2) &&\in \mathbb{R}^{H, d_h} \\
a_t &= exp(-softplus(h_t \cdot W_A)) &&\in \mathbb{R}^{H} \\
(q^c_t)^{\otimes2} &= q^c_t \otimes q^c_t &&\in \mathbb{R}^{H, D} \\
(k^c_t)^{\otimes2} &= k^c_t \otimes k^c_t &&\in \mathbb{R}^{H, D} \\
H^\uparrow_t &= H^\uparrow_{t-1} \odot a_t + \left[ (k^c_t)^{\otimes2} \right]^T \cdot (v^c_t) &&\in \mathbb{R}^{H, D, d_h} \\
H^\downarrow_t &= H^\downarrow_{t-1} \odot a_t + \left[ (k^c_t)^{\otimes2} \right]^T &&\in \mathbb{R}^{H, D} \\
y_N &= \frac{(q^c_t)^{\otimes2} \cdot H^\uparrow_t}{(q^c_t)^{\otimes2} \cdot H^\downarrow_t} &&\in \mathbb{R}^{H, d_h} \\
o_t &= y_t \cdot W_{out} &&\in \mathbb{R}^{d} \\[-4pt]
\end{align*}
\end{algorithmic}
\end{algorithm}
\section{Model Ablation Details}
In this section, we briefly mention the setup for all ablations. Specifically, the llama 2 (\citet{llama2}) architecture is used as our base model. To ablate the changes to the attention mechanism, we replace the attention mechanism in all layers of the llama 2 model with the ablated variations, keeping all other parts of the architecture constant. The base max sequence length is 2048, however 4096 and 8192 sequence lengths are used for testing models on longer context lengths. Most experiments are done with a base model of about 300 million parameters. To test scale we increase the size to 700 million parameters. The following is information regarding our training setup:
\begin{itemize}
    \item dataset: HuggingFaceFW/fineweb (CC-MAIN-2024-51 version)
    \item test dataset: held out 0.1 percent of the entire dataset (seed 123)
    \item model name: meta-llama/Llama-2-7b-hf
    \item optimizer: AdamW (0.9, 0.999 betas)
    \item batch size (over all devices): 32
    \item learning rate: 1e-4
    \item warmup steps: 10,000
    \item total steps: 100,000
    \item weight decay: 0.01
    \item no grad clipping
\end{itemize}
The 300 million parameter model (small) has the following setup:
\begin{itemize}
    \item hidden size: 1024
    \item intermediate MLP size: 2048
    \item hidden MLP activation: silu
    \item num attention heads: 16 (64 head dim)
    \item num layers: 20
    \item vocab size: 32,000 (llama 2 vocab)
\end{itemize}
The 700 million parameter model (medium) has the following setup:
\begin{itemize}
    \item hidden size: 1536
    \item intermediate MLP size: 3072
    \item hidden MLP activation: silu
    \item num attention heads: 24 (64 head dim)
    \item num layers: 27
    \item vocab size: 32,000 (llama 2 vocab)
\end{itemize}
\section{Gradients}
\label{app:gradients}
We derive the gradient equations for all necessary kernels and provide them in this section. Each gradient assumes the input is the query, key, and value matrices and the output is the post-attention outputs. Each kernel performs the attention operation, but no pre-processing or post-processing.
\subsection{Linear}
\begin{align*}
    O &= \left( Q K^T \odot M \right) V \\
    \frac{\partial L}{\partial Q} &= \left( \frac{\partial L}{\partial O} V^T \odot M \right) K \\
    \frac{\partial L}{\partial K} &= \left(V \frac{\partial L}{\partial O}^T \odot M^T \right) Q \\
    \frac{\partial L}{\partial V} &= \left( Q K^T \odot M \right)^T \frac{\partial L}{\partial O}
\end{align*}
\subsection{Linear with Softmax Norm}
\begin{align*}
    O &= \frac{ Q K^T \odot M }{ \sum \left( Q K^T \odot M \right) } V \\
    Y &= Q K^T \odot M \\
    S &= \sum_j Y \\
    Y_N &= Norm(Y) = \frac{Y}{N} = \frac{Y}{\sum Y} \\
    G &= \frac{\partial L}{\partial O} V^T \odot M \\
    D &= \frac{G - \sum_j \left( Y_N \odot G \right)}{S} \odot M \\
    \frac{\partial L}{\partial Q} &= D K \\
    \frac{\partial L}{\partial K} &= D^T Q \\
    \frac{\partial L}{\partial V} &= Y_N^T \frac{\partial L}{\partial O}
\end{align*}
\subsection{Linear with A-Gate}
Note: Typically we use $M$ to denote the causal mask. Here, we additionally use M to denote the row (key and value) axis whereas the column (query) axis is denoted by $N$. Additionally, $\sum_j$ denotes a summation along the $M$ axis while $\sum_i$ denotes a summation along the $N$ axis. The differentiation is necessary for the gradients of the $A$-mask. In the self-attention case, $N=M$, however we still make this differentiation for the gradient derivation below.
\begin{align*}
    A_M &= e^{A - A^T} \quad \in \mathbb{R}^{B, H, N, M}, \quad A \in \mathbb{R}^{B, H, N} \\
    O &= \left( Q K^T \odot M \odot A_M \right) V \\
    \frac{\partial L}{\partial Q} &= \left( \frac{\partial L}{\partial O} V^T \odot M \odot A_M \right) K \\
    \frac{\partial L}{\partial K} &= \left( \frac{\partial L}{\partial O} V^T \odot M \odot A_M \right)^T Q \\
    \frac{\partial L}{\partial V} &= \left( Q K^T \odot M \odot A_M \right)^T \frac{\partial L}{\partial O} \\
    \frac{\partial L}{\partial A^N_M} &= \sum_j \left( Q K^T \odot M \odot A_M \odot \frac{\partial L}{\partial O} V^T \right) \\
    \frac{\partial L}{\partial A^M_M} &= -\sum_i \left( Q K^T \odot M \odot A_M \odot \frac{\partial L}{\partial O} V^T \right) \\
    \frac{\partial L}{\partial A_M} &= \frac{\partial L}{\partial A^N_M} + \frac{\partial L}{\partial A^M_M}
\end{align*}
\subsection{Squared with A-mask}
\begin{align*}
     A_M &= e^{A - A^T} \quad \in \mathbb{R}^{B, H, N, M}, \quad A \in \mathbb{R}^{B, H, N} \\
    O &= \left( (Q K^T)^2 \odot M \odot A_M \right) V \\
    \frac{\partial L}{\partial Q} &= \left( 2 \odot Q K^T \odot \frac{\partial L}{\partial O} V^T \odot M \odot A_M \right) K \\
    \frac{\partial L}{\partial K} &= \left( 2 \odot Q K^T \odot \frac{\partial L}{\partial O} V^T \odot M \odot A_M \right)^T Q \\
    \frac{\partial L}{\partial V} &= \left( (Q K^T)^2 \odot M \odot A_M \right)^T \frac{\partial L}{\partial O} \\
    \frac{\partial L}{\partial A^N_M} &= \sum_j \left( (Q K^T)^2 \odot M \odot A_M \odot \frac{\partial L}{\partial O} V^T \right) \\
    \frac{\partial L}{\partial A^M_M} &= -\sum_i \left( (Q K^T)^2 \odot M \odot A_M \odot \frac{\partial L}{\partial O} V^T \right) \\
    \frac{\partial L}{\partial A_M} &= \frac{\partial L}{\partial A^N_M} + \frac{\partial L}{\partial A^M_M}
\end{align*}
\subsection{Squared with A-mask and Softmax Norm (\AlgoName{})}
\begin{equation*}
\begin{split}
    O &= \frac{ \left( Q K^T \right)^2 \odot M \odot A_M }{ \sum \left( \left( Q K^T \right)^2 \odot M \odot A_M \right) } V \\
    Y &= \left( Q K^T \right)^2 \odot M \odot A_M \\
    S &= \sum_j Y \\
    Y_N &= Norm(Y) = \frac{Y}{S} \\
    G &= V \frac{\partial L}{\partial O}^T \odot A_M \odot M - \sum_j \left( Y_N \odot \frac{\partial L}{\partial O} V^T \right) \\
    D &= \frac{2 \odot QK^T \odot G}{S} \odot M \\
    \end{split}
\qquad \qquad 
    \begin{split}
    \frac{\partial L}{\partial Q} &= D K \\
    \frac{\partial L}{\partial K} &= D^T Q \\
    \frac{\partial L}{\partial V} &= Y_N^T \frac{\partial L}{\partial O} \\
    \frac{\partial L}{\partial A^N_M} &= \sum_j \left( Y_N \odot G \right) \\
    \frac{\partial L}{\partial A^M_M} &= -\sum_i \left( Y_N \odot G \right) \\
    \frac{\partial L}{\partial A_M} &= \frac{\partial L}{\partial A^N_M} + \frac{\partial L}{\partial A^M_M}
\end{split}
\end{equation*}
\subsection{Exponentiated with A-mask and Softmax Norm (\AlgoNameExp{})}
\begin{equation*}
\begin{split}
    O &= \frac{ e^{Q K^T} \odot M \odot A_M }{ \sum \left( e^{Q K^T} \odot M \odot A_M \right) } V \\
    Y &= e^{Q K^T} \odot M \odot A_M \\
    S &= \sum_j Y \\
    Y_N &= Norm(Y) = \frac{Y}{S} \\
    G &= \frac{\partial L}{\partial O} V^T \odot M - \sum_{d_h} \left( O \odot \frac{\partial L}{\partial O} \right) \\
    D &= Y_N \odot G \odot M \\
    \end{split}
\qquad \qquad 
    \begin{split}
    \frac{\partial L}{\partial Q} &= D K \\
    \frac{\partial L}{\partial K} &= D^T Q \\
    \frac{\partial L}{\partial V} &= Y_N^T \frac{\partial L}{\partial O} \\
    \frac{\partial L}{\partial A^N_M} &= \sum_j D \\
    \frac{\partial L}{\partial A^M_M} &= -\sum_i D \\
    \frac{\partial L}{\partial A_M} &= \frac{\partial L}{\partial A^N_M} + \frac{\partial L}{\partial A^M_M}
    \end{split}
\end{equation*}
\section{Pile and SlimPJ Loss Curves}
To verify \AlgoName{} generalizes to other datasets, we perform a training run with medium sized models on The Pile (\citet{pile}) and SlimPajama (\citet{SlimPajama}).
\begin{figure}[htbp]
    \centering
    \begin{subfigure}[b]{0.48\textwidth}
        \centering
        \includegraphics[width=\linewidth]{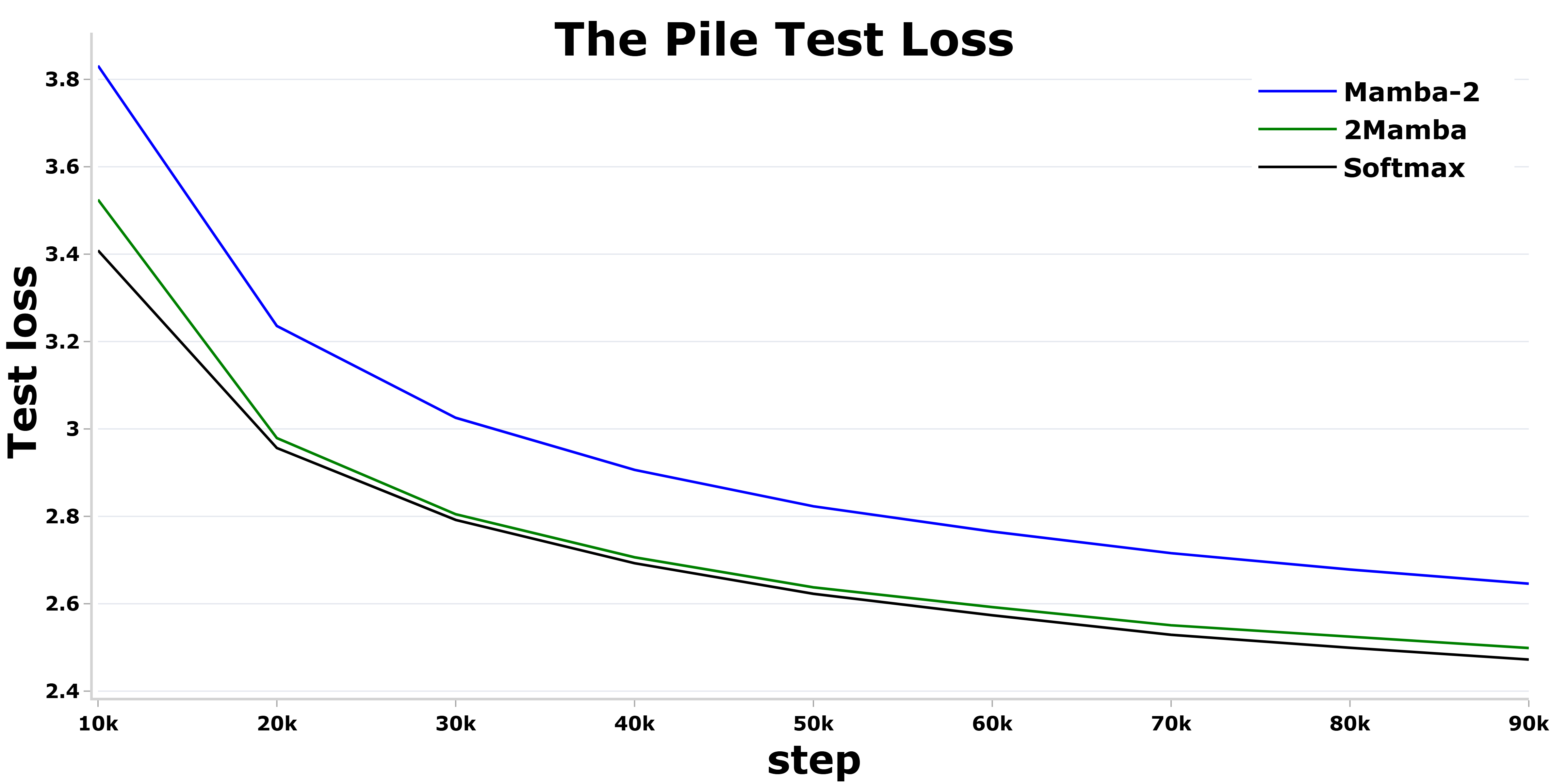}
        \caption{Test loss on The Pile}
        \label{fig:pile}
    \end{subfigure}
    \hfill
    \begin{subfigure}[b]{0.48\textwidth}
        \centering
        \includegraphics[width=\linewidth]{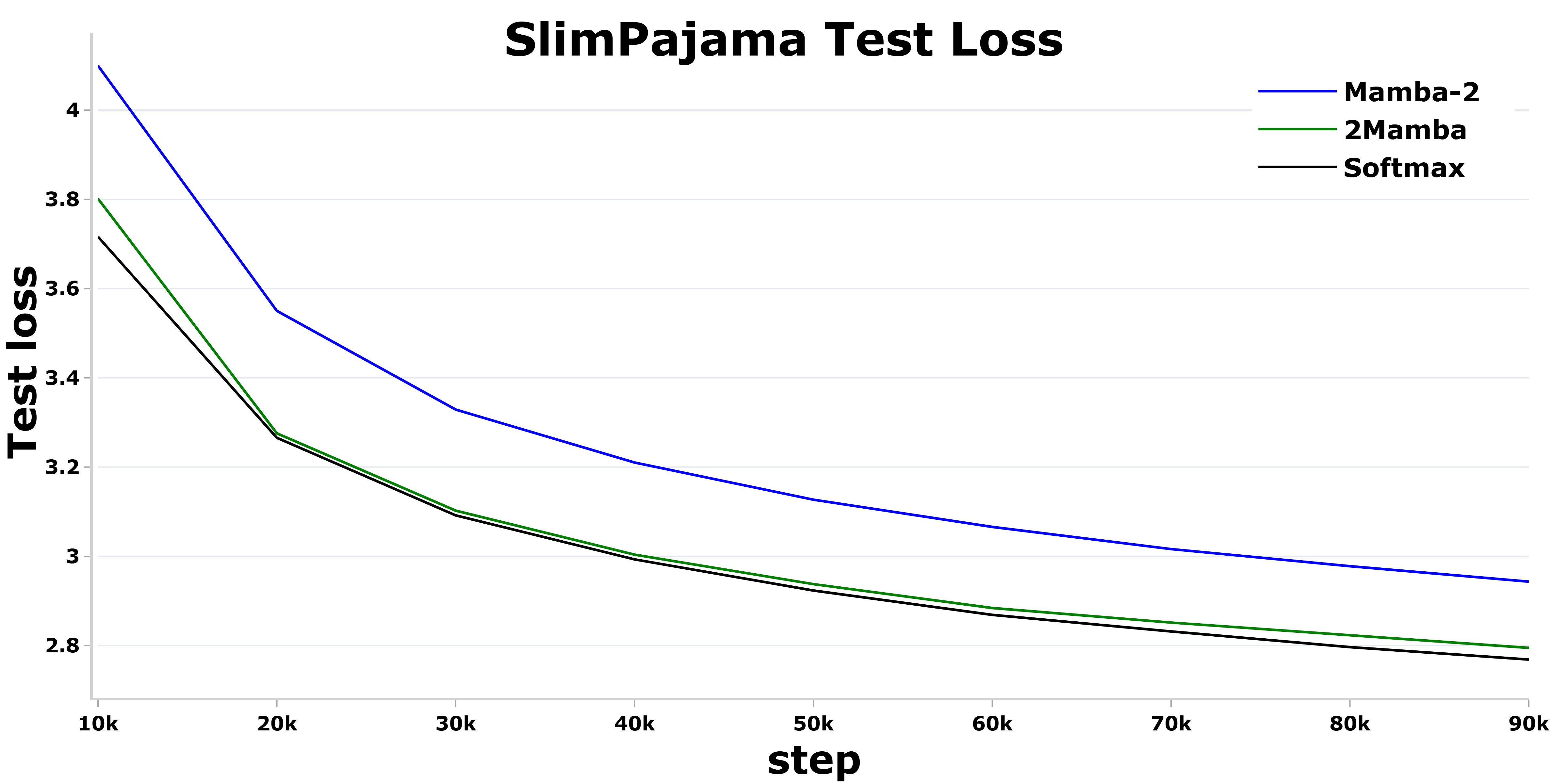}
        \caption{Test loss on SlimPajama}
        \label{fig:slimpj}
    \end{subfigure}
\end{figure}
\end{document}